%% file: 00_main.tex
\documentclass{egpubl}
\usepackage{pg2023}

\SpecialIssuePaper         % uncomment for final version of Computer Graphics Forum, special issue
\CGFStandardLicense
\usepackage[T1]{fontenc}
\usepackage{dfadobe}  
\usepackage{cite}  % comment out for biblatex with backend=biber
\BibtexOrBiblatex
\electronicVersion
\PrintedOrElectronic
\ifpdf \usepackage[pdftex]{graphicx} \pdfcompresslevel=9
\else \usepackage[dvips]{graphicx} \fi

\usepackage{egweblnk} 
\usepackage{amsfonts}
\usepackage{amsmath}
\usepackage{multirow}
\usepackage{booktabs}
\usepackage[caption=false]{subfig}
\usepackage[dvipsnames]{xcolor}
\renewcommand\footnotemark{}
\title[IBL-NeRF]{IBL-NeRF: Image-Based Lighting Formulation of\\Neural Radiance Fields}

\author[Choi et al.]
{\parbox{\textwidth}{\centering Changwoon Choi$^*$\phantom{\thanks{* Authors contributed equally to this work.\\ \hphantom{\scriptsize{...}}$\dag$ Corresponding author.}}\hspace{-0.5em}$^{1}$\orcid{0000-0001-5748-6003} Juhyeon Kim$^{*}$$^{1,2}$\orcid{0000-0002-6218-3426} Young Min Kim$^\dag$
$^{1}$\orcid{0000-0002-6735-8539}
}
\\
{\parbox{\textwidth}{\centering $^1$Department of Electrical and Computer Engineering, Seoul National University\\
$^2$Department of Computer Science, Dartmouth College
       }
}
}
\renewcommand\footnotemark{\arabic{footnote}}

\begin{document}

\teaser{
 \includegraphics[width=\linewidth]{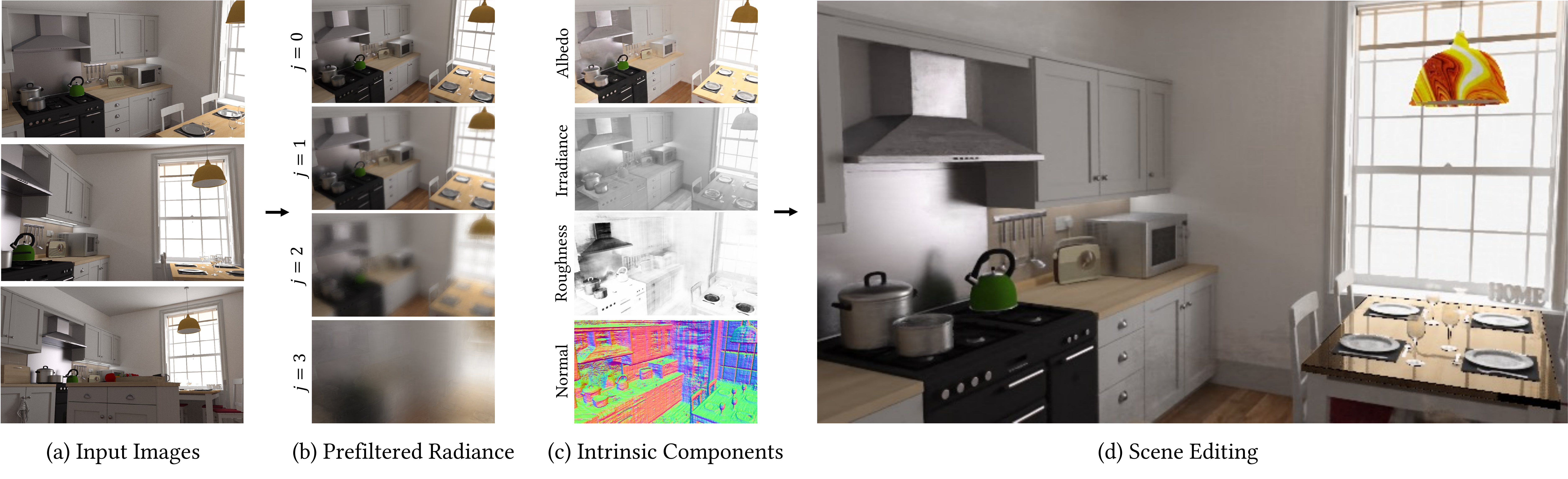}
 \centering
  \caption{We propose IBL-NeRF, a neural volume representation with prefiltered radiance field inspired by image-based lighting formulation. (a) Given multi-view images, we optimize the (b) prefiltered radiance field and estimate (c) reflectance properties of the material (albedo, roughness), lighting information (irradiance, prefiltered radiance), and the geometry (normal). (d) One can manipulate the neural scene easily by modifying the decomposed components.
  Project page: \url{https://changwoon.info/publications/IBL-NeRF}}
\label{fig:teaser}
}

\maketitle
%-------------------------------------------------------------------------
\begin{abstract}
    We propose IBL-NeRF, which decomposes the neural radiance fields (NeRF) of large-scale indoor scenes into intrinsic components.
    % Previous approaches for the intrinsic decomposition of NeRF transform the implicit volume to fit the rendering pipeline of explicit geometry, and approximate the views of segmented, isolated objects with environment lighting.
    % Previous approaches for the intrinsic decomposition of NeRF replace baked color with a rendering equation, and they estimate the intrinsic components and lighting that make up the rendering equation.
    Recent approaches further decompose the baked radiance of the implicit volume into intrinsic components such that one can partially approximate the rendering equation.
    % However, they are limited to representing isolated objects and suffer from computational burden since they use a single environment lighting for the whole scene with Monte Carlo integration.
    However, they are limited to representing isolated objects with a shared environment lighting, and suffer from computational burden to aggregate rays with Monte Carlo integration.
    In contrast, our prefiltered radiance field extends the original NeRF formulation to capture the spatial variation of lighting within the scene volume, in addition to surface properties.
    Specifically, the scenes of diverse materials are decomposed into intrinsic components for rendering, namely, albedo, roughness, surface normal, irradiance, and prefiltered radiance.
    All of the components are inferred as neural images from MLP, which can model large-scale general scenes.
    Especially the prefiltered radiance effectively models the volumetric light field, and captures spatial variation beyond a single environment light.
    The prefiltering aggregates rays in a set of predefined neighborhood sizes such that we can replace the costly Monte Carlo integration of global illumination with a simple query from a neural image.
    By adopting NeRF, our approach inherits superior visual quality and multi-view consistency for synthesized images as well as the intrinsic components.
    We demonstrate the performance on scenes with complex object layouts and light configurations, which could not be processed in any of the previous works.
%-------------------------------------------------------------------------
%  ACM CCS 1998
%  (see http://www.acm.org/about/class/1998)
% \begin{classification} % according to http:http://www.acm.org/about/class/1998
% \CCScat{Computer Graphics}{I.3.3}{Picture/Image Generation}{Line and curve generation}
% \end{classification}
%-------------------------------------------------------------------------
%  ACM CCS 2012
% (see http://www.acm.org/about/class/class/2012)
%The tool at \url{http://dl.acm.org/ccs.cfm} can be used to generate
% CCS codes.
%Example:



\end{abstract}  
\input{01_introduction}
\input{02_related_works}

\input{03_methods}

\input{04_experiments}
\input{05_conclusion}

\paragraph*{Acknowledgements}
This work was partly supported by Korea Institute for Advancement of Technology (KIAT) grant funded by the Korea Government (MOTIE) (P0012746, HRD Program for Industrial Innovation), and the BK21 FOUR program of the Education and Research Program for Future ICT Pioneers, Seoul National University in 2023.

\bibliographystyle{eg-alpha-doi}
\bibliography{99_references}

\appendix
\input{11_BRDF}

\input{12_prefiltered_radiance}

\input{13_additinal_results}

\end{document}

%% file: 01_introduction.tex
\section{Introduction}

% simplicity of formulation of NeRF
Neural radiance field (NeRF) ~\cite{mildenhall2020nerf} prospers for their superior quality in novel-view synthesis with a simple formulation. 
A neural network is trained to optimize a colored density volume to directly match multiple posed input images. 
The formulation is ignorant of any intermediate representations of traditional rendering pipelines, namely surface geometry, light transport, or BRDF. 
%Due to the generality of the capturing set-up, it can efficiently represent complex scenes with non-local lighting effects or complex material.
The trained volumetric representation does not trace iterative inter-reflections of rays, or model complex occlusion of the surface geometries.
Nonetheless, NeRF can produce detailed subtleties of global illumination and parallax effects.

% intrinsic decomposition and NeRF
While NeRF can capture complex effects in general scenes, the implicit formulation limits further analysis or edits of the scenes. 
Intrinsic decomposition is an attractive choice as it decomposes the captured scene into intrinsic components that can be further manipulated to edit the scene.
However, intrinsic decomposition is inherently an ill-posed problem and requires enforcing additional priors or constraints.
Prior works often extract an isolated object in a bounding box, selected with exhaustive segmentation masks, for intrinsic decomposition of NeRF. 
They assume a single low-dimensional environment lighting for the entire scene and incorporate additional knowledge for reflectance properties, such as priors on BRDFs or images captured under different known illuminations.
% Under the constrained set-up, they partially transform the segmented objects into forward rendering with Monte-Carlo integration which can be computationally expensive.
% \jkim{(r3)what does partially transform mean?} 
% \ccw{Under the constrained set-up, they transform the segmented objects into forward rendering with Monte-Carlo integration which can be computationally expensive.}
% \ykim{Under the constrained set-up, they sample rays between the approximated environment light and the segmented object with Monte-Carlo integration which can be computationally expensive.}
Under the constrained set-up, they sample rays between the approximated environment light and the segmented object with Monte-Carlo integration which can be computationally expensive.
Furthermore, such approximation with environment light prohibits viewpoints inside the scene, or a local variation of lights caused from common light fixtures or windows.
By relinquishing the flexibility of the original NeRF, existing inverse rendering with NeRF approaches cannot represent everyday environments composed of diverse unsegmented objects.

% description of IBL-NeRF
Instead of extensively simulating multiple bounces of rays with approximated explicit representation, we propose incorporating constraints from the image spaces, extending the NeRF formulation.
Specifically, we train a decomposed neural volume, coined IBL-NeRF, to optimize for the implicit light distribution of neural images.
This neural representation captures detailed spatial variations of lighting, in contrast to low-dimensional environment mapping. 
Then we can substitute the illumination integration process into a simple network query for the irradiance.
The specular reflection of different surface roughness values is fetched from prefiltered radiance fields of appropriate prefilter levels, similar to texture mipmap.
We additionally enforce priors on the intrinsic components for input images, acquired from existing methods for decomposing individual images. 
By incorporating image-based lighting with implicit intrinsic components, we can efficiently render general scenes without sacrificing the rendering quality of the original NeRF as shown in Fig.~\ref{fig:teaser}. 
We can further edit scenes by changing materials or adding objects, including highly reflective or transparent objects.

In summary, our approach fully leverages the high-quality novel view images of the original NeRF formulation, and yet enables efficient re-generation with approximations inspired from image-based lighting. 
Our contributions can be listed as following:
\begin{itemize}
 % generalizable, flexible set-up
 \item We propose IBL-NeRF, which handles global illuminations with spatially varying lighting and diverse materials given a set of unsegmented images.
 % efficiency of formulation
 \item We model the prefiltered radiance of the scene with a neural network of NeRF, and efficiently approximate rendering equations with image-based lighting formulation.
 % scene edits are possible
 \item  Our neural representation extracts physically interpretable components of the complex indoor scenes which can be altered to render images with different attributes.
\end{itemize} 
\vspace{-0.5em}
The results are presented with large-scale scenes containing multiple objects, which can not be modeled with previous works employing a single environment lighting with Monte-Carlo integration.
\vspace{-0.5em}

%% file: 02_related_works.tex
\section{Related Works}

While NeRF~\cite{mildenhall2020nerf} can synthesize photo-realistic novel-view images, one of its limitations is that the radiance information is baked within the implicit neural representation.
Several subsequent works propose to distill intrinsic components, such as illumination and reflectance property, and try to achieve inverse rendering with implicit representation, in contrast to reconstructing explicit mesh geometry with multi-view stereo~\cite{10.1145/3469842, duchene2015multi}.
% They optimize components to match the input images by adopting Monte Carlo (MC) integration, which requires heavy computation.
They optimize components to match the input images by evaluating the rendering equation with Monte Carlo (MC) method, which requires heavy computation.
% ~\jkim{(r3) by evaluating rendering equation with MC method}
Neural Reflectance Fields~\cite{bi2020reflectance_fields} and NeRV~\cite{srinivasan2021nerv} adapt ray-marching to account for reflectance, and model the illumination with a single point light and environment light, respectively.
% \jkim{(r3) change order? (seems wrong)}
% \ccw{I think that we were right.}
Both approaches require multiple images with known lighting configurations as input.
NeRFactor~\cite{zhang2021nerfactor}, Hasselgren et al.~\cite{hasselgren2022nvdiffrecmc}, NeRD~\cite{boss2021nerd}, and PhySG~\cite{zhang2021physg}, on the other hand, factorize radiance fields from unknown light.
They concurrently optimize for a single low-dimensional environment light in a coarse resolution (NeRFactor, Hasselgren et al.) or spherical Gaussian (NeRD, PhySG).

In contrast, IBL-NeRF proposes to efficiently synthesize images without explicit Monte-Carlo integration, and utilizes prefiltered radiance which can be evaluated with a single ray sample.
Several works~\cite{verbin2021refnerf, boss2021neuralpil} also adapt integrated illumination for efficient rendering.
They are either implicitly conditioned on the surface reflectance property, or propose components without physical interpretation.
However, previous works using integrated illumination employ a single environment lighting for entire scene and therefore are limited to modeling an isolated object.
Concurrent works, such as I$^2$-SDF~\cite{zhu2023i2} and TexIR~\cite{li2022texir}, also exploit spatially-varying light for complex indoor scenes, but they use MC integration or explicit mesh representation, respectively.

% \jkim{(r3)Inverse rendering to Intrinsic scene reconstruction or relightable scene reconstruction}
% Inverse rendering for general scenes requires modeling spatially-varying lighting.
% \ccw{Intrinsic decomposition for general scenes requires modeling spatially-varying lighting.}
Intrinsic decomposition for general scenes requires modeling spatially-varying lighting.
With increased degrees of freedom for the already under-constrained problem, scene decomposition requires strong assumptions.
Commonly used priors include piece-wise constant albedos~\cite{cheng2018intrinsic, li2018cgintrinsics, li2018learning, ma2018single, laffont2012coherent}, or sparsity of extracted albedo values~\cite{meka2021real, garces2012intrinsic}.
A few works exploit data-driven priors instead of hand-crafted priors~\cite{bell2014intrinsic, zhou2015learning, sengupta2019neural, li2020inverse, 10.1145/3450626.3459872, yi2020leveraging}, which can be subject to domain discrepancy.
% A few works exploit data-driven priors instead of hand-crafted priors~\cite{bell2014intrinsic, zhou2015learning, sengupta2019neural, li2020inverse, 10.1145/3450626.3459872}.
IBL-NeRF takes inspiration from the aforementioned prior works using single images, and adds constraints in the image space.
Because the neural volume of NeRF is trained with images, the formulation can readily be applied to handle challenging indoor scenes without simplifying the illumination model.
Furthermore, IBL-NeRF can naturally find multi-view consistent components, which is not possible with single-image decomposition.

%% file: 03_methods.tex
\newcommand{\norm}[1]{\left\lVert#1\right\rVert}
\newcommand{\Lprefj}{L^j_{\text{pref}}(\mathbf{x},\omega)}
\newcommand{\xsurf}{\mathbf{x}_{\text{surf}}}
\newcommand{\Lprefjr}{L^j_{\text{pref}}(\xsurf,\omega_r)}
\newcommand{\Lprefjno}{L^j_{\text{pref}}}
\newcommand{\Lprefjnotilde}{\tilde{L}^j_{\text{pref}}}
\newcommand{\Lprefjnohat}{\hat{L}^j_{\text{pref}}}

\begin{figure*}[t!]
    \centering
    \includegraphics[width=\linewidth]{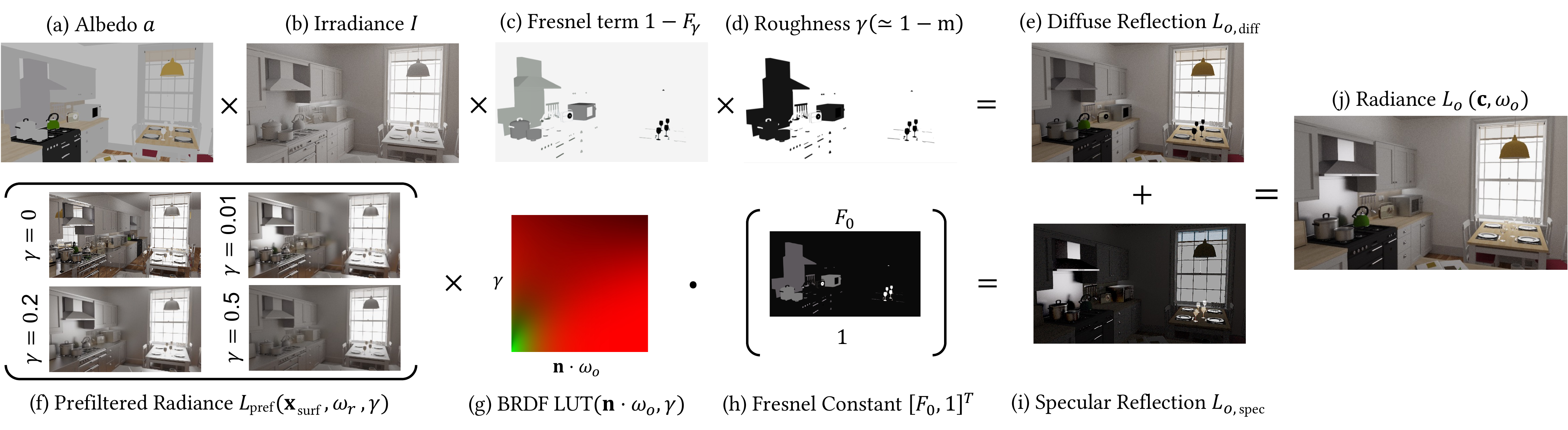}
    \caption{Overview of the radiance approximation used in IBL-NeRF (Eq.~\ref{eq:approximated_rendering_eq}). With the combination of inferred (a) albedo, (b) Irradiance, (c) Fresnel term, and (d) roughness, one can obtain diffuse reflection. Also, with (f) multi-level prefiltered radiance, (g) fetched value from LUT and (h) Fresnel constant, we can calculate (i) specular reflection. (j) Final approximated radiance is achieved by the sum of diffuse and specular reflection.}
    % \jkim{(g) $n -> \mathbf{n}$}\ccw{$\rightarrow$ fixed!}}
    \label{fig:IBL-NeRF_overview}
\end{figure*}
\section{Method}

\subsection{IBL-NeRF Formulation}

\subsubsection{Preliminaries}
\label{subsec:preliminaries}

Ray-tracing engines approximate the light transport with samples of rays. %, which is computationally expensive.
The original rendering equation~\cite{kajiya1986rendering} formulates the outgoing radiance at surface $\xsurf$ as a combination of reflected rays of incoming radiance $L_i$
\begin{equation}
L_o(\xsurf, \omega_o)=\int_\Omega f_r(\xsurf,\omega_i,\omega_o)L_i(\xsurf,\omega_i)(\mathbf{n}\cdot\omega_i)d\omega_i,
    \label{eq:rendering_eq}
\end{equation}
where $\mathbf{n}$ and $f_r$ are the surface normal and BRDF at surface $\xsurf$, and $\omega_i$ and $\omega_o$ are incoming and outgoing direction.
Given the scene properties ($\mathbf{n}$ and $f_r$), the rendered output relies on the diverse distribution of light transport, $L_i$ and $L_o$, which are 5D functions.

The approximation within game engines~\cite{karis2013real} replaces the recursive calls of radiances $L_i \rightarrow L_o$ into a single sample of integrated light.
$L_o$  is approximated as the sum of two components, namely the diffuse term and the specular term:
% \begin{align} \label{eq:approximated_rendering_eq}
%     L_o(\mathbf{x},\omega_o) &= \underbrace{\gamma \times (1-F_\gamma (\omega_o, \mathbf{n}, \gamma)) \times a \times I}_{L_{o,\text{diff}}} \nonumber \\
%     & + \underbrace{L_{\text{pref}}(\mathbf{x}, \omega_r,\gamma)\times \left[F_0,1\right]^T \text{LUT}(\omega_o\cdot\mathbf{n},\gamma)}_{L_{o,\text{spec}}}.
% \end{align}
\begin{multline} \label{eq:approximated_rendering_eq}
    L_o(\xsurf,\omega_o) = \underbrace{\gamma \times (1-F_\gamma (\omega_o, \mathbf{n}, \gamma)) \times a \times I}_{L_{o,\text{diff}}} \\
    + \underbrace{L_{\text{pref}}(\xsurf, \omega_r,\gamma)\times \left[F_0,1\right]^T \text{LUT}(\omega_o\cdot\mathbf{n},\gamma)}_{L_{o,\text{spec}}}.
\end{multline}
% \jkim{(r3) $I -> I(\xsurf, \mathbf{n})$}
The diffuse term depends on irradiance $I=\int_{\Omega}L_i(\xsurf,\omega_i)(\mathbf{n}\cdot\omega_i)d\omega_i$ which integrates all the incoming radiance.
Additionally, it is proportional to the surface albedo $a$, roughness $\gamma$, and approximated Fresnel term $F_\gamma$. 
(According to the original paper of \cite{karis2013real}, the diffuse term is attenuated by (1-metallic) and we approximated it as roughness ($\gamma$). % Further approximation could be tried in future works.
More sophisticated approximations could be tried in future works.)
% (We approximate the 1-metallic in \cite{karis2013real} to roughness.)
Calculating the specular term $L_{o,\text{spec}}$ involves directional components of rays.
The split-sum approximation simplifies the specular term into the product of two terms.
The first component $L_{\text{pref}}$ is the \textit{prefiltered environment map} which summarizes the effects of reflected lights to mimic specular highlights efficiently.
It is filtered according to the surface roughness level $\gamma$ and fetched at the reflected direction.
The second component is also precalculated as a 2D lookup texture (LUT).
Both diffuse term and specular term are affected by roughness $\gamma$.
IBL-NeRF allows the decomposition of NeRF by utilizing neural network to represent the pre-computed volumetric light distribution.
Detailed descriptions of the approximation are available in the supplementary material. 
% Detailed descriptions of the approximation are available in the Appendix \ref{sec:BRDF}.

\begin{figure}
    \centering
    \includegraphics[width=0.7\linewidth]{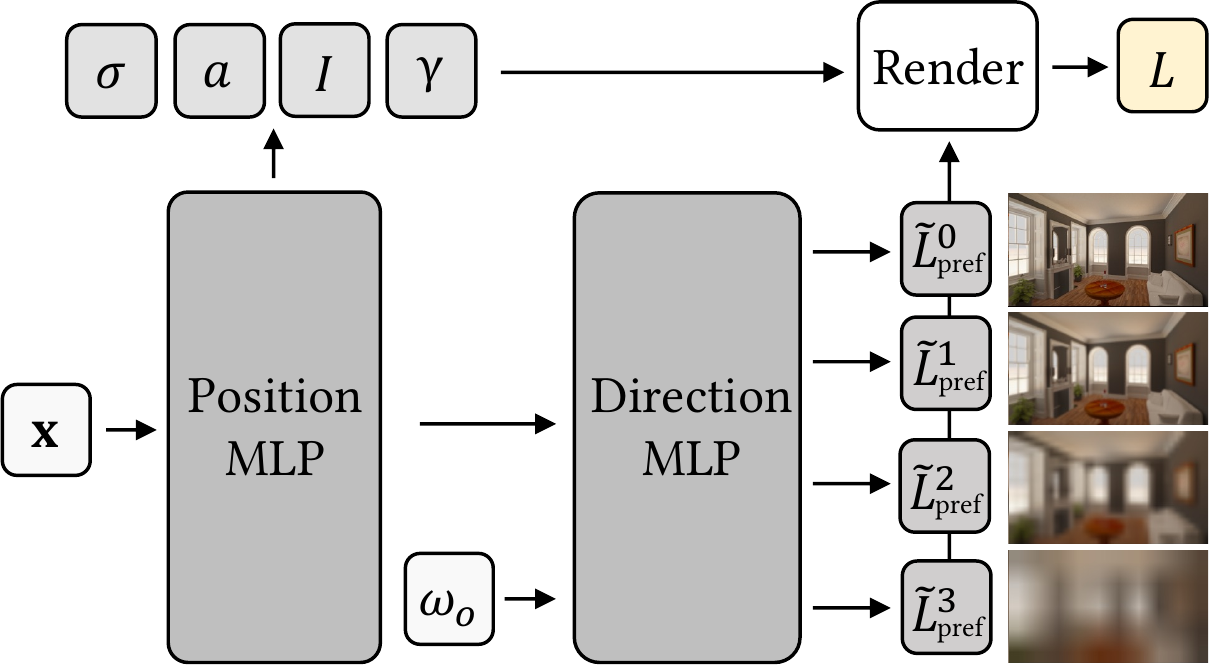}
    \caption{Architecture of IBL-NeRF. The scene properties dependent on position (volume density $\sigma$, albedo $a$, irradiance $I$ and roughness $\gamma$) are extracted from position MLP. Those dependent on the viewing direction (each level of prefiltered radiance $\tilde{L}_{\text{pref}}^j$) are obtained from direction MLP.}
    \label{fig:architecture}
\end{figure}

\subsubsection{Rendering Pipeline of IBL-NeRF}
\label{subsubsec:rendering_IBL-NeRF}

NeRF synthesizes a photo-realistic image applying a volume rendering on a neural volume
\begin{equation}
    L_o (\mathbf{c}, \omega_o) = \int_0^{\infty}V(\mathbf{x}(t), \mathbf{c})\sigma (\mathbf{x}(t))L_e(\mathbf{x}(t), \omega_o)dt,
    \label{eq:volume_rendering}
\end{equation} 
where $\mathbf{x}(t)=\mathbf{c}-t\omega_o$ represents points on a ray initiated from the camera position $\mathbf{c}$, and $V(\mathbf{x}(t),\mathbf{c})=\exp{\left(-\int_0^t \sigma (\mathbf{x}(s))ds\right)}$ is the visibility.
Given a position $\mathbf{x}$ and an outgoing direction $\omega_o$, the neural volume of NeRF is trained to regress for density $\sigma$ from the positional MLP and the emitted radiance $L_e$ from the directional MLP.
The training objective is to match the results of volume rendering with the pixels in the input images, which enables creating images of the scene only from a set of multiple-view images.

To decompose the radiance of NeRF into physically interpretable components of the scene, we can adapt components as presented in Eq.~\ref{eq:approximated_rendering_eq}, %Sec.~\ref{subsec:preliminaries},
ignorant of light transport.
For each ray, we evaluate albedo $a$, irradiance $I$, and roughness $\gamma$ with the volume density $\sigma$.
We accumulate the values along the ray using volume rendering following the NeRF formulation in Eq.~\ref{eq:volume_rendering}.
Also, at the estimated surface point, the network evaluates the prefiltered radiance field $L_{\text{pref}}$ of the reflected direction.
Due to the computational complexity, the reflected rays are evaluated only at the surface hit position of the ray, which is estimated as $\mathbf{x}_{\text{surf}}=\mathbf{c}-d\omega_o$~\cite{zhang2021nerfactor, srinivasan2021nerv}.
The termination depth $d(\mathbf{c}, -\omega_o)$ of the ray defines the surface point $\mathbf{x}_{\text{surf}}$ and can be obtained with density $\int_0^\infty \exp{\left(-\int_0^\infty \sigma (\mathbf{c}-s\omega_o)ds \right)t\sigma (\mathbf{c} - t\omega_o)}dt$.
% \begin{equation}
%     \int_0^\infty \exp{\left(-\int_0^\infty \sigma (\mathbf{c}-s\omega_o)ds \right)t\sigma (\mathbf{c} - t\omega_o)}dt.
% \end{equation}
We obtain the surface normal from the numerical gradient of the termination depth $d$: ($\mathbf{n}(\mathbf{x}_{\text{surf}}) = \nabla_\mathbf{x} d(\mathbf{x},\omega)/\lVert \nabla_\mathbf{x} d(\mathbf{x},\omega )\rVert$.)
All the values are combined using Eq.~\ref{eq:approximated_rendering_eq} to find the output radiance corresponding to the pixel, which is also visualized in Fig.~\ref{fig:IBL-NeRF_overview}.

Fig.~\ref{fig:architecture} shows the modified neural network architecture.
The positional MLP infers the components that do not have view dependency, namely, albedo $a$, irradiance $I$, and roughness $\gamma$, in addition to the volume density $\sigma$ in the vanilla NeRF.
% \jkim{(r3) Note that irradiance is inferred in an implicit way, not explicitly integrating over the hemisphere.}
% \ccw{Note that irradiance is inferred from MLP implicitly, not explicitly integrating over the hemisphere.}
% \ykim{Note that irradiance is inferred from MLP implicitly, instead of explicitly integrating over the hemisphere.}
Note that irradiance is inferred from MLP implicitly, instead of explicitly integrating over the hemisphere.
The irradiance depends on surface normal, but we assume that it is implicitly handled in the neural network, which takes position $\mathbf{x}$ as input.
The directional component is encoded as prefiltered radiance field $L_{\text{pref}}$, and is the output of the subsequent directional MLP. 
It is modulated by roughness $\gamma$ and combined to generate the final image. 
The following subsection further explains the formulation and approximation used for the prefiltered radiance fields.

\definecolor{myblue}{RGB}{183, 199, 218}
\definecolor{myyellow}{RGB}{241, 204, 177}

\begin{figure}
    \centering
    \includegraphics[width=\linewidth]{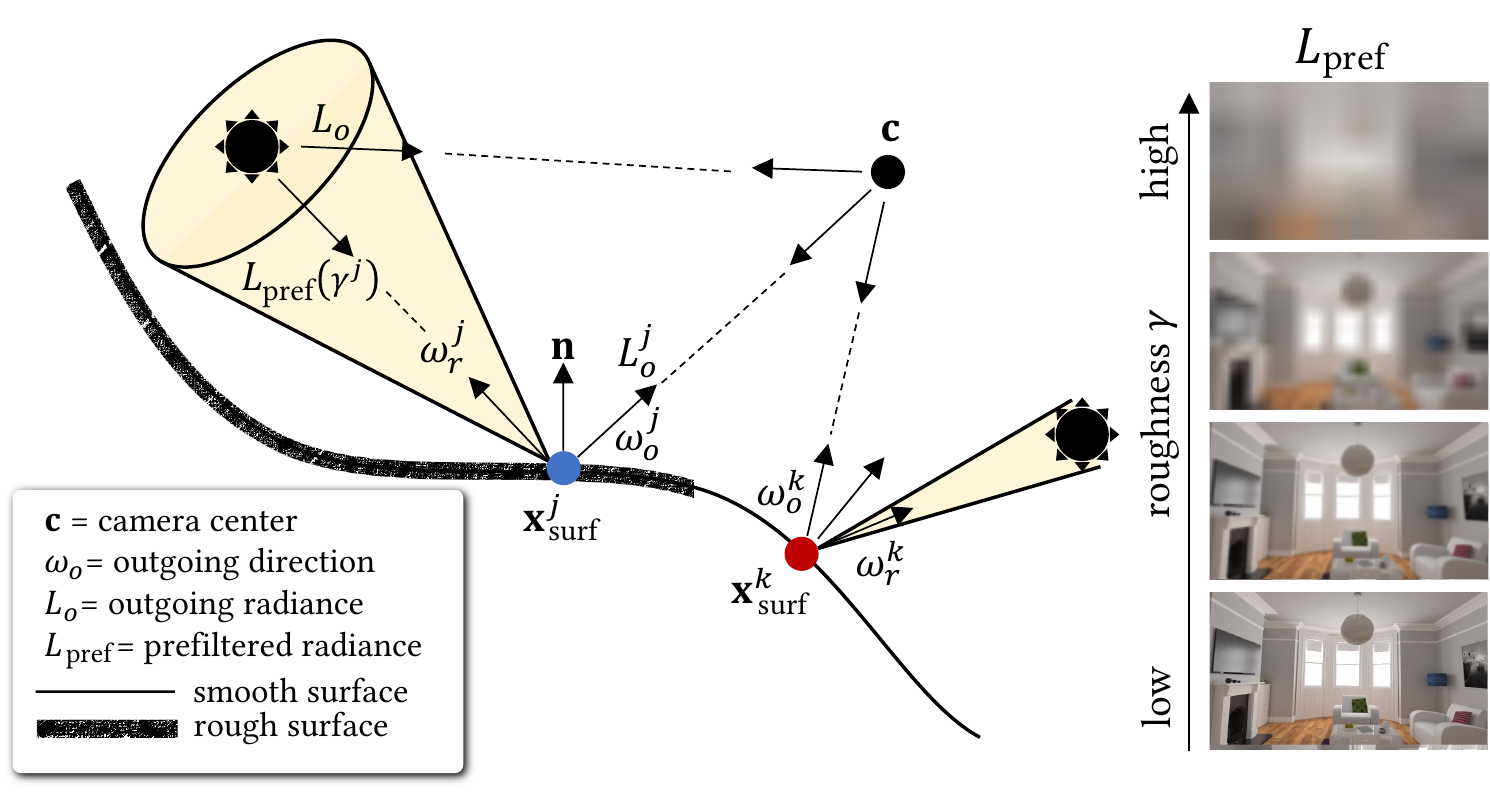}
    \caption{Specular reflection of IBL-NeRF. The prefiltered radiance field is fetched only from the estimated surface point $\mathbf{x}_{\text{surf}}$ with a single reflected ray toward the direction of mirror reflection $\omega_r$. The point on rough surface (intersection of $j^{\text{th}}$ ray) fetches the prefiltered radiance convoluted with a wide kernel. In contrast, the point on smooth surface (intersection of $k^{\text{th}}$ ray) reads prefiltered radiance field filtered with a narrow kernel.}
    \label{fig:IBL-NeRF_explain}
\end{figure}

\subsection{Prefiltered Radiance Fields}

The prefiltered environment map $L_{\text{pref}}(\xsurf, \omega_r,\gamma)$ in Eq.~\ref{eq:approximated_rendering_eq} accounts for the specular reflection with directional components that reside in a high-dimensional space as a single sample.
Let us denote the camera observation direction as $-\omega_o$  and its mirror reflection with respect to the surface normal $\mathbf{n}$ at the surface point as $\omega_r$.
Unless the surface is a perfect mirror (roughness $0$), the reflected rays are evaluated within angular distribution near the reflection direction.
As the surface roughness $\gamma$ increases, prefiltered radiance should be filtered with a wider range kernel.
Fig.~\ref{fig:IBL-NeRF_explain} illustrates the procedure, where the pre-filtered radiance at $\omega_r$ is depicted with cones with yellow shade, whose angle indicates the size of convolution kernel for the roughness value.

\begin{figure}
    \centering
    \includegraphics[width=0.9\linewidth]{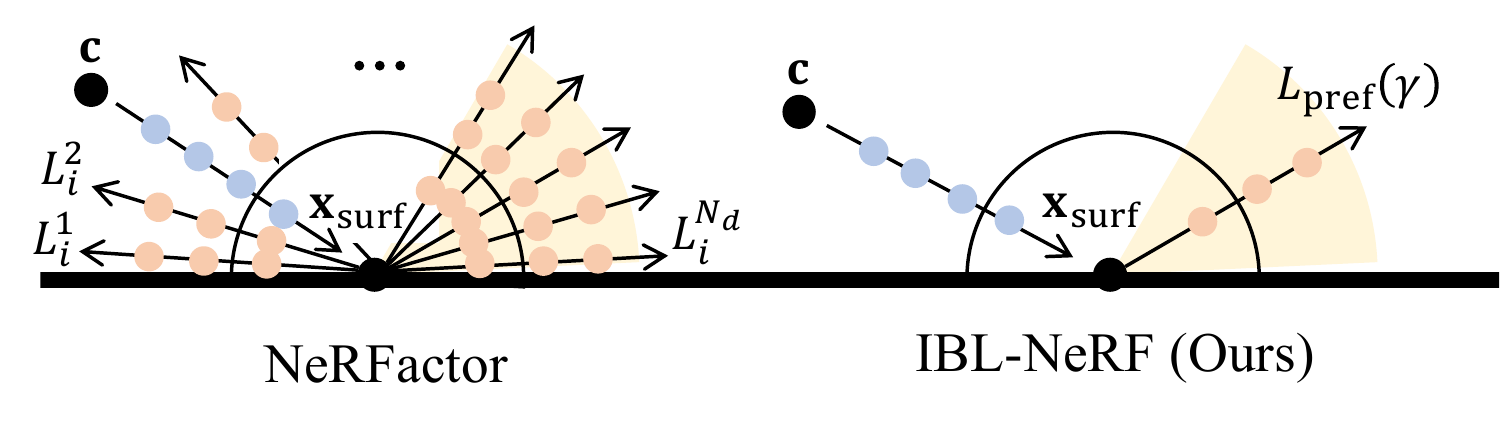}
    \caption{
    %Supplementary visual material for the time 
    Illustration for the time complexity analysis in Table~\ref{tab:method_comparison}. Prefiltered radiance fields significantly reduce the sample number during the training phase from $\mathcal{O}(\textcolor{myblue}{N_s}+\textcolor{myyellow}{N_d N_r})$ to $\mathcal{O}(\textcolor{myblue}{N_s}+\textcolor{myyellow}{N_r})$.
    % \jkim{NeRFactor -> full hemisphere}
    % \ccw{fixed}
    }
    \label{fig:method_compare_graphic}
\end{figure}

While there exist several works that approximate specular illumination from a hit point, IBL-NeRF alleviates the need for Monte-Carlo integration and greatly reduces the computational burden.
Table~\ref{tab:method_comparison} summarizes the comparison of IBL-NeRF against NeRFactor~\cite{zhang2021nerfactor}, which is a representative formulation with environment light~\cite{zhang2021nerfactor, srinivasan2021nerv, bi2020reflectance_fields}.
Specifically, the Monte-Carlo integration aggregates $N_d$ directional samples of reflected rays from the surface points as shown in shaded cones in Fig.~\ref{fig:IBL-NeRF_explain}.
In addition to the $N_s$ samples along the camera ray for the volume rendering of NeRF, each reflected ray is evaluated with $N_r$ samples of towards the surrounding environment lighting.
Although NeRFactor directly fetch $N_d$ light samples from environment map according to the visibility MLP output for each direction, they need to query $N_r$ samples along each direction to train visibility MLP which is originally used in NeRV~\cite{srinivasan2021nerv}.
% \jkim{(r3)which is originally used in NeRV~\cite{srinivasan2021nerv}}.
% \jkim{why inside ()?}
% \ykim{we could remove the parenthesis!}
The variants using Monte-Carlo integration therefore require evaluating $\mathcal{O}(N_s + N_d N_r)$ samples.
On the other hand, IBL-NeRF proposes fetching a single ray of the prefiltered radiance field $L_{\text{pref}}(\mathbf{x}_\text{surf},\omega_r,\gamma)$ in the place of the environment map, leading to evaluating $\mathcal{O}(N_s + N_r)$ samples, as depicted in Fig.~\ref{fig:method_compare_graphic}.

\input{tables/method_comparison}

Additionally, IBL-NeRF can process general scenes with diverse lighting or viewpoints as long as the original NeRF converges.
The prefiltered radiance fields is defined for the entire scene volume for any position $\mathbf{x}_\text{surf}$ and or direction $\omega_r$.
This is in contrast to the approaches relying on a single environment light which is an infinite-radius spherical image enclosing the entire scene, as they assume an isolated object distant from other scene properties, especially lighting.
Therefore it cannot render from viewpoints within the volume, diverse objects spread throughout the scene, or indoor scenes with interior lighting.

Our specular reflection is evaluated as a single ray for the given roughness value within the scene volume since the prefiltered radiance $L_{\text{pref}}(\xsurf, \omega_r,\gamma)$ already aggregates the directional rays.
Specifically, IBL-NeRF outputs prefiltered radiance fields $L_{\text{pref}}^j$ with different convolution levels $j$.
The prefiltered radiance of the desired roughness $\gamma$ at a certain point $\mathbf{x}$ with direction $\omega$ uses  trilinear interpolation as
\begin{equation}
    L_{\text{pref}} (\mathbf{x},\omega, \gamma ) = \sum_j w^j(\gamma)L_{\text{pref}}^j(\mathbf{x}, \omega),
    \label{eq:trilinar_interpolation}
\end{equation}
where $w^j(\gamma)$ is the weight of $j$th mipmap that depends on the roughness $\gamma$ as described in Fig.~\ref{fig:IBL-NeRF_explain}.
%\diff{The weight of prefiltered radiance fields $w^j(\gamma)$ is obtained by linear interpolation using roughness value.}
The values stored in the prefiltered radiance fields correspond to specific roughness values, and we linearly interpolate them to adjust to the current $\gamma$ value.
Therefore, we evaluate the prefiltered radiance by fetching a sample of a single ray, similar to texture mipmap.

The prefiltered radiance $\Lprefjnotilde$ is inferred from the directional MLPs using the similar volume rendering equation
\begin{equation}
    \Lprefjno(\textbf{c}, - \omega_o) = \int_{0}^{\infty} V(\textbf{x}(t), \mathbf{c})\sigma(\textbf{x}(t)) \Lprefjnotilde(\textbf{x}(t), - \omega_o) dt.
\end{equation}
For training, we use a set of images blurred with a discrete set of Gaussian filters from the camera position $-\omega_o$.
During the inference of the image, the values of $\Lprefjnotilde$ are fetched to render the surface point $\mathbf{x}_{\text{surf}}$ as explained in Sec.~\ref{subsubsec:rendering_IBL-NeRF} and Eq.~\ref{eq:trilinar_interpolation}.
Note that the training target is the blurred images observed from the camera $(\mathbf{c}, -\omega_o)$, whereas the inference is evaluated from the reflected direction $(\mathbf{x}_{\text{surf}}, \omega_r)$.
The formulation relies on the assumption that training images contain observations of the reflected rays.

\subsubsection{Image-Space Approximation}

The prefiltered radiance $L_{\text{pref}}^j$ of IBL-NeRF incorporates the image-based rendering within the implicit volume of NeRF and achieves computational efficiency.
We further analyze the practical considerations with the image-space approximation of Gaussian filters to emulate the specular reflection blobs of different surface roughness.
The $j$th prefiltered radiance $L_\mathrm{pref}^j$ is approximated for the roughness value $\gamma_j$ as
\begin{align}
    L_{\text{pref}}^j % & = \mathbb{E}_{\omega_i\sim p(\omega_i|\mathbf{x}, \omega, \gamma_j)} [L_i(\mathbf{x}, \omega_i)] \label{eq:pref_j_E}\\
    & = \int_\Omega L_i(\mathbf{x}, \omega_i) p(\omega_i|\mathbf{x}, \omega, \gamma_j) d \omega_i \label{eq:pref_j_omega}\\
    & = \int_S L_i(s_i) p_S(s_i|\mathbf{x}, \omega, \gamma_j) d s. \label{eq:pref_j_screen} 
\end{align}
Previous approaches approximate the sampling distribution by inferring radiance multiple times in hemispherical domain $\Omega$ (Eq.~\ref{eq:pref_j_omega}) which is computationally heavy~\cite{zhang2021nerfactor, srinivasan2021nerv}.
Our method converts the domain into the image space $S$ of the current view as Eq.~\ref{eq:pref_j_screen},
where $s_i$ is the screen space coordinate that corresponds to direction $\omega_i$.
When rendering for a viewpoint, the viewing direction $\omega_o$ can be assumed to be constant, and we can use a globally consistent kernel $\Lprefj=K^j(L(s))$, where  $K^j(s_i) \propto p_S(s_i|\mathbf{x},\omega,\gamma_j)$.
% We include the full derivation of our approximation and plots of $K^j(s_i)$ in the Appendix~\ref{appendixb}.
We include the full derivation of our approximation and plots of $K^j(s_i)$ in the supplementary material.
The overall shape of $K^j(s_i)$ is similar to that of the Gaussian function, which is used to approximate $\Lprefj$ in our implementation.
%\diff{Note that assuming using a perspective camera, our image-space approximation deviates from environment map filtering more as the image pixel approaches the depth edge.}
While Gaussian kernels in image space mostly result in a reasonable approximation, they deviate from direct filtering of the environment map for pixels near the image edge.
Additional discussion on our approximation can be found in the supplementary material.
% Additional discussion on our approximation can be found in the Appendix~\ref{appendixb}.

\newcommand{\Lprefno}{L_{\text{pref}}}

\subsection{Training IBL-NeRF}

IBL-NeRF imposes constraints on the rendered images to train the neural volume, similar to vanilla NeRF.
The objective function is composed of four terms:
\begin{equation}
    \mathcal{L} = \mathcal{L}_{\text{render}} + \mathcal{L}_{\text{pref}}  +\mathcal{L}_{\text{prior}} + \lambda_{I, \text{reg}}\mathcal{L}_{I,\text{reg}}.
\end{equation}
The first two components are rendering losses to match the rendered images with the input images.
For each pixel of the camera ray $r=(\mathbf{c},-\omega_o)$, the rendering loss $\mathcal{L}_\text{render}$ of approximated radiance is defined as 
\begin{equation}
    \mathcal{L}_{\text{render}} = \lVert L_o(r) - \hat{L}_o(r) \rVert_2^2,
\end{equation}
where $\hat{L}_o$ is ground truth radiance and ${L}_o$ is our approximated radiance calculated with Eq.~\ref{eq:approximated_rendering_eq}.
$\mathcal{L}_{\text{pref}}$ is the rendering loss of prefiltered radiance defined as
\begin{equation}
    \mathcal{L}_{\text{pref}} = \sum_j \lVert {L}^j_{\text{pref}}(r) - L^j_{\text{G}}(r) \rVert_2^2.
\end{equation}
$L^j_{\text{pref}}$ is inferred prefiltered radiance of $j^{\text{th}}$ level and $L^j_{\text{G}}$ is the radiance convolved with $j^{\text{th}}$ level Gaussian convolution, where $L^0_{\text{G}}=L$.

Inverse rendering is under-constrained in nature, and the remaining two losses incorporate additional prior knowledge to estimate intrinsic components.
We obtain the pseudo albedo $\hat{a}$ and irradiance $\hat{I}$ for our input images by applying intrinsic decomposition for single images~\cite{bell2014intrinsic}, and use them as data-driven prior.
The prior loss $\mathcal{L}_{\text{prior}}$ encourages our inferred albedo $a$ to match the pseudo albedo
\begin{equation}
    \mathcal{L}_{\text{prior}} = \lVert a(r) - \hat{a}(r) \rVert_2^2.
\end{equation}
In addition, $\mathcal{L}_{I,\text{reg}}$ is the irradiance regularization loss 
\begin{equation}
    \mathcal{L}_{I,\text{reg}} = \lVert I(r) - \mathbb{E}[\hat{I}] \rVert_2^2,
\end{equation}
where $\mathbb{E}[\hat{I}]$ is the mean of irradiance (shading) values in training set images.
 Although the results from single-image decomposition are inconsistent for different viewpoints, our neural volume learns multi-view consistent and smooth results.
% Supplementary material contains more detailed comparison between IBL-NeRF and results from single-image decomposition methods.
We provide more detailed comparison between IBL-NeRF and results from single-image decomposition methods in Sec.~\ref{subsec:experiments_view_synthesis_and_intrinsic_decomposition}.

%% file: tables/method_comparison.tex
\begin{table}
\centering
\resizebox{\linewidth}{!}{
\begin{tabular}{ccccc} 
    \toprule
  {} & NeRF & NeRFactor & IBL-NeRF (Ours) \\
    \midrule
 Rendering & Volume &Surface & Surface\\ 
 $L_o$ & Baked & Monte Carlo Integration & Approx w. Eq. ~\ref{eq:approximated_rendering_eq} \\ 
 $L_i$ & - & Env light w. Visibility Infer & MLP Inference\\ 
 \begin{tabular}{@{}c@{}}Time \\ Complexity \end{tabular} & $\mathcal{O}(N_s)$ & $\mathcal{O}(N_s+N_d N_r)$&$\mathcal{O}(N_s+N_r)$ \\
 \bottomrule
 
\end{tabular}
}
%\\ \includegraphics[width=0.8\linewidth]{figures/time_complex_graphic.pdf}
\caption{We compare IBL-NeRF with NeRF and a recent method decomposing NeRF's radiance.
The time complexity is measured for the entire training phase.
% \jkim{(r3) emphasize it is training phase}
% \ccw{I think we explained sufficiently}
$N_s$ and $N_r$ are the numbers of samples along a camera ray and a reflected ray, and $N_d$ is the number of directional samples over a hemisphere.}
\label{tab:method_comparison}
\end{table}

%% file: 04_experiments.tex
\section{Experiments}

\begin{figure}
    \centering
    \includegraphics[width=0.95\linewidth]{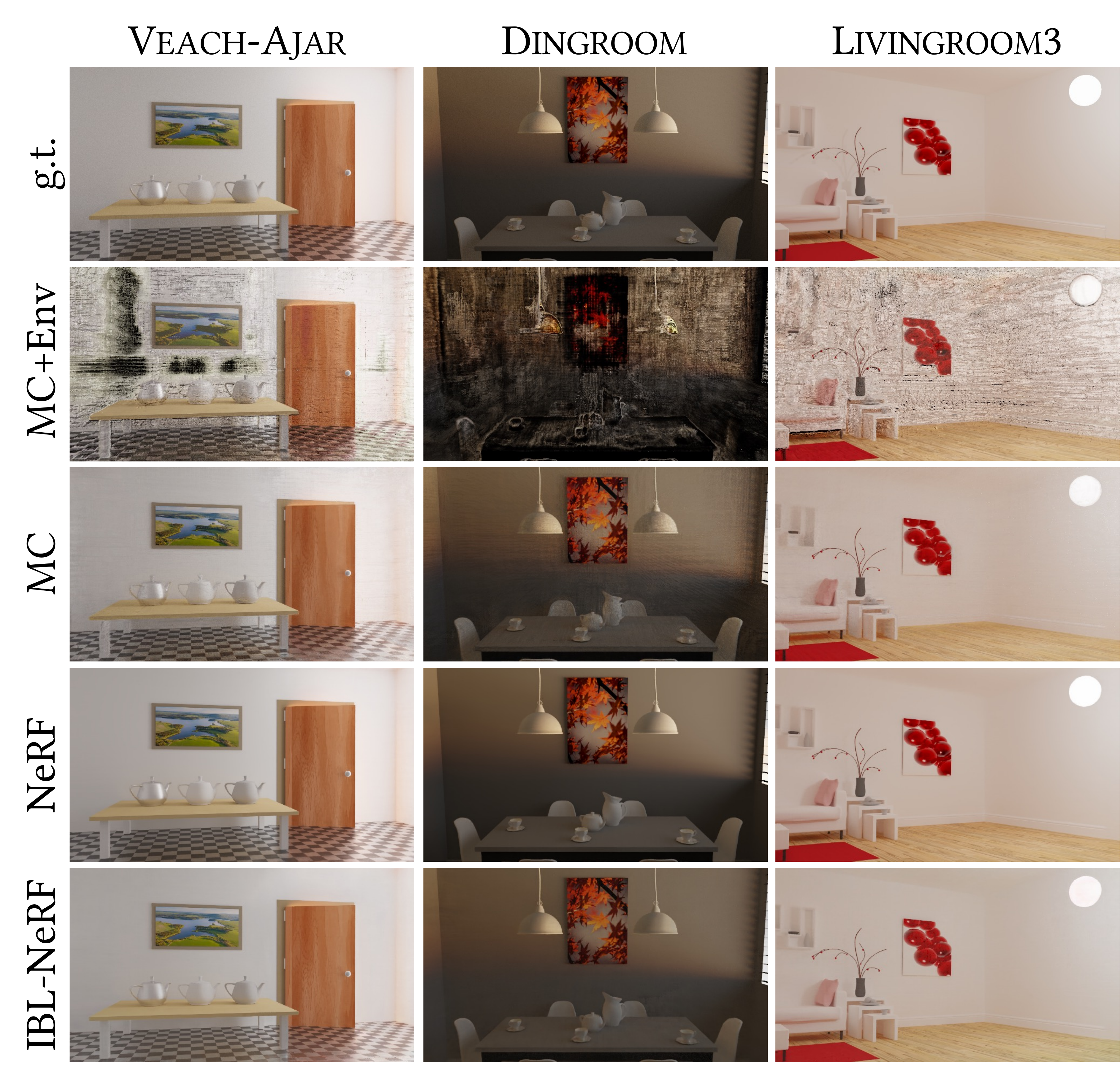}
    \caption{Qualitative results of novel-view image synthesis.}
    \label{fig:view_synthesis}
\end{figure}
\input{tables/view_synthesis}

\noindent\textbf{Dataset}
First, we test IBL-NeRF in 12 realistic synthetic indoor scenes~\cite{resources16}, which are capable of obtaining ground-truth intrinsic components.
We render 100 multi-view images for both training and test set with the OptiX~\cite{parker2010optix} based path tracer~\cite{10.2312:pg.20211379}.
All of the scenes in our dataset exhibit complex lighting with windows or interior lighting and contain multiple objects with challenging material, which cannot be modeled with an environment light.
This is in contrast to previous works for decomposing NeRF, which present results with isolated objects~\cite{zhang2021nerfactor,srinivasan2021nerv}.
The camera’s position and rotation are randomly sampled within a scene bounding box. All the results reported in the manuscript are the novel viewpoints in the testset which are not seen during the training images.
We linearly interpolate between the test camera poses to generate results of the supplementary video.
Furthermore, we test IBL-NeRF in real-world scenes from ScanNet dataset~\cite{dai2017scannet} and our own captured scene.
For ScanNet scenes, we use train/test split from~\cite{Wei_2021_ICCV}.
The camera poses are estimated with COLMAP~\cite{Schonberger_2016_CVPR} for real scenes.

\noindent\textbf{Implementation Details}
The neural network architecture is illustrated in Fig.~\ref{fig:architecture}.
We use the same MLP configurations with vanilla NeRF~\cite{mildenhall2020nerf}, except that IBL-NeRF has additional layers to output albedo ($a$), irradiance ($I$), and roughness ($\gamma$) at position MLP, and 4 parallel layers to emit the prefilterd radiance fields ($L_\text{pref}^j$) for each roughness level $0\leq j\leq3$.
IBL-NeRF is trained for 120k steps with 512 ray samples, and follows the training schedule below to stabilize the process.
For the first 10k steps, we only optimize $L_\text{pref}^j$ and $\sigma$ with $ \mathcal{L_{\text{pref}}}$.
Once we obtain stable geometry and prefiltered radiance fields, we additionally optimize for $\mathcal{L}_\text{pref} + \mathcal{L}_\text{render}$ without prior.
Then we freeze roughness and apply priors $\mathcal{L}_{\text{prior}}, \mathcal{L}_{I,\text{reg}}$ for last 20k steps.
We use $\lambda_{I,\text{reg}}=0.1$ empirically, but we observe that the final result is not very sensitive to the value of $\lambda_{I,\text{reg}}$.
Also, we assume monochromatic irradiance for simplicity.
% \jkim{(r2) Discuss HDR? (did we use HDR, tonemapping?)}

\begin{figure}
    \centering
    \includegraphics[width=0.85\linewidth]{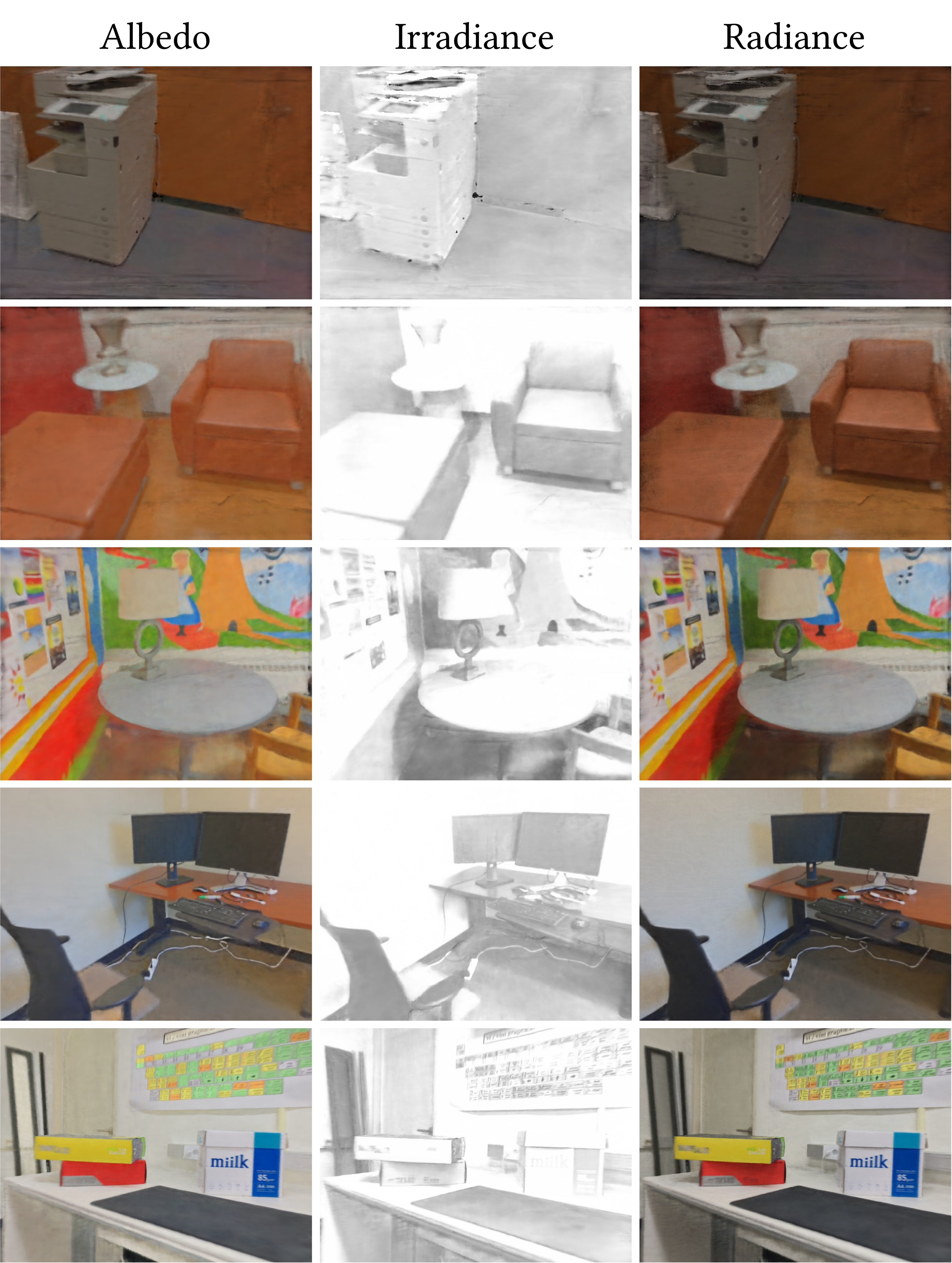}
    \caption{Qualitative results of intrinsic decomposition and view synthesis on real-world datasets.}
    \label{fig:intrinsic_decomposition_real}
\end{figure}
\input{tables/intrinsic_decomposition_only}

\subsection{View Synthesis \& Intrinsic Decomposition}
\label{subsec:experiments_view_synthesis_and_intrinsic_decomposition}

% \begin{figure*}
%     \centering
%     \includegraphics[width=\linewidth]{figures/intrinsic_decomposition_v2.pdf}
%     \caption{Qualitative results of intrinsic decomposition and view synthesis on our synthetic datasets.}
%     \label{fig:intrinsic_decomposition}
% \end{figure*}

\begin{figure*}
    \centering
    \includegraphics[width=\linewidth]{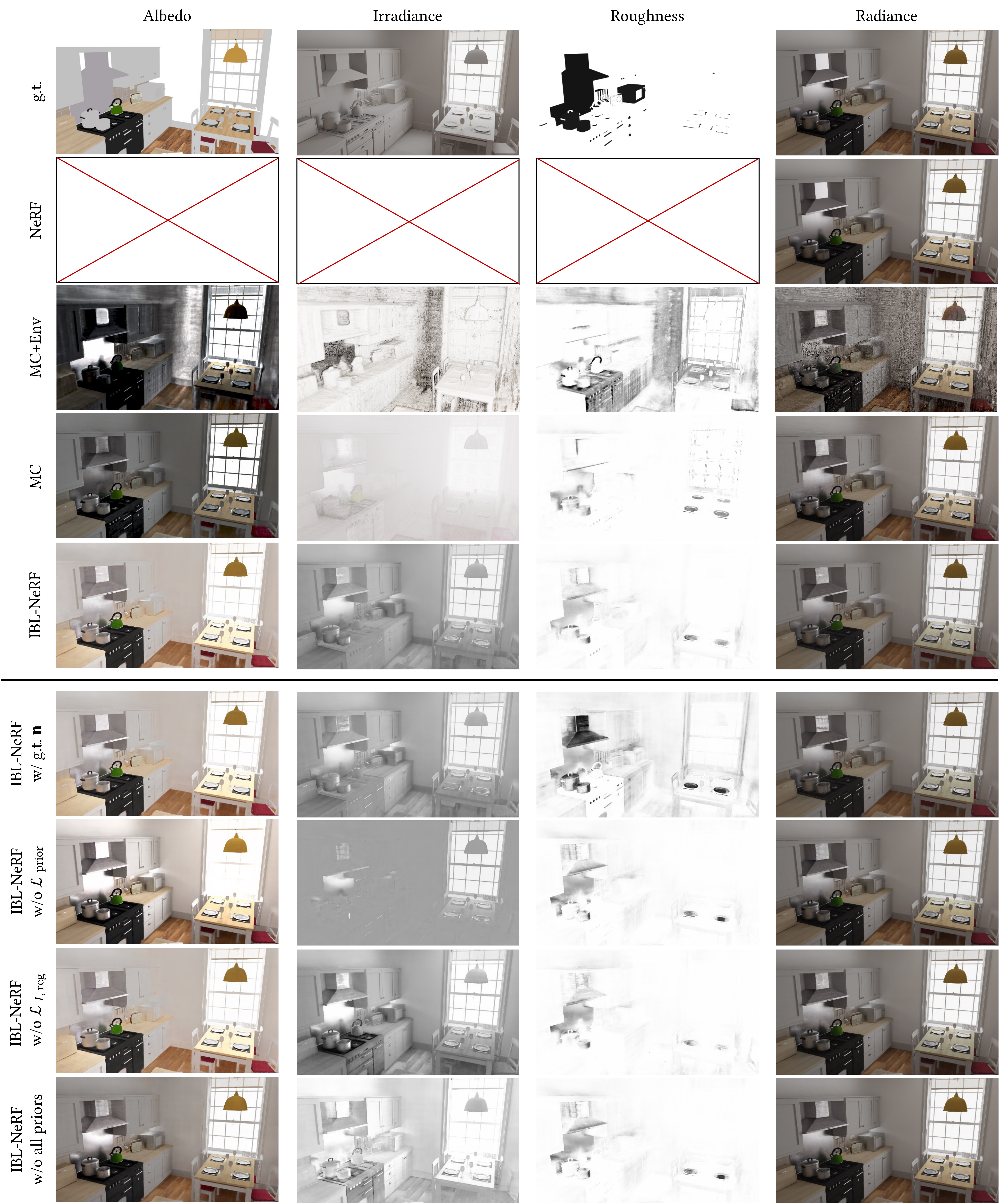}
    \caption{Qualitative results of intrinsic decomposition and view synthesis on \textsc{Kitchen}.}
    \label{fig:intrinsic_decomposition_kitchen}
\end{figure*}

\begin{figure*}
    \centering
    \includegraphics[width=\linewidth]{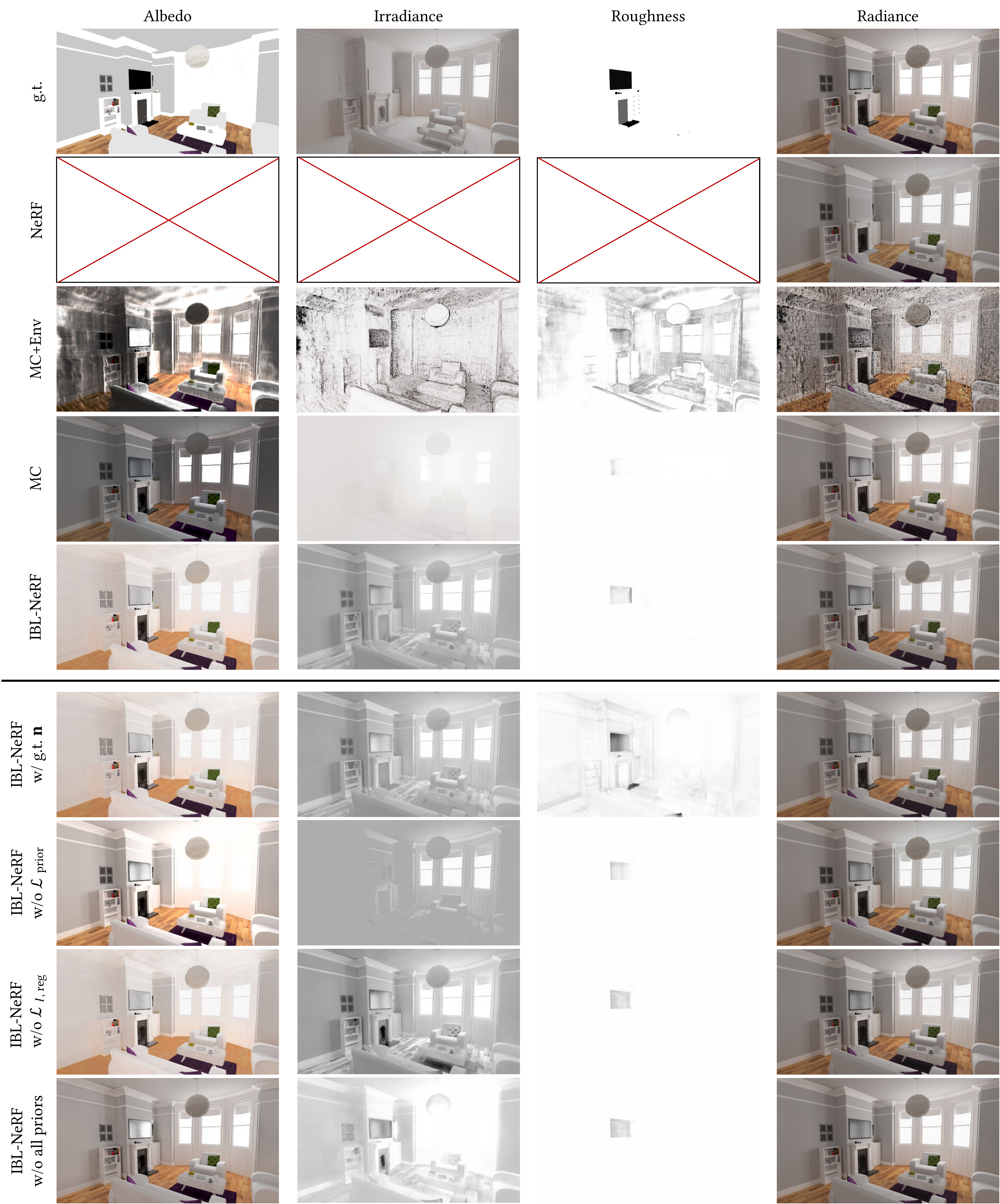}
    \caption{Qualitative results of intrinsic decomposition and view synthesis on \textsc{Livingroom2}.}
    \label{fig:intrinsic_decomposition_livingroom}
\end{figure*}

\noindent\textbf{Baselines}
We compare IBL-NeRF with two baselines with Monte Carlo (MC) sampling over a hemisphere of environment light.
Since Neural Reflectance Fields~\cite{bi2020reflectance_fields} and NeRV~\cite{srinivasan2021nerv} need known lighting formulation to train models, they cannot be applied to our scenario with unknown lighting conditions.
% The first baseline (MC) is a variant of IBL-NeRF, which estimates prefiltered radiance $L_\text{pref}$ and calculates integration with MC sampling.
The first baseline (MC) is a variant of IBL-NeRF, which exploits the radiance field ($L_\text{pref}^0=L_o$) as incoming light for specular reflection and calculates integration with MC sampling.
The second baseline (MC + Env) estimates single environment light for the entire scene as $L_i$ and employs MC integration as in NeRFactor and NeRV.
% \jkim{(r2) The environment map is also trained as NeRFactor or NeRV.}
% MC + Env is the microfacet BRDF version of NeRFactor~\cite{zhang2021nerfactor} which is the most relevant work to us.
% \ccw{MC + Env is the microfacet BRDF version of NeRFactor~\cite{zhang2021nerfactor} without visibility infer network which is the most relevant work to us.}
% \jkim{(r3) MC + Env is NeRFactor without visibility infer network.}
% \ykim{MC + Env is the microfacet BRDF version of NeRFactor~\cite{zhang2021nerfactor} without visibility inference network, which is the most relevant work to us.}
MC + Env is the microfacet BRDF version of NeRFactor~\cite{zhang2021nerfactor} without visibility inference network, which is the most relevant work to us.
We found that the original NeRFactor which exploits learned BRDF prior does not converge in any of our scenes.
For all the baselines with Monte Carlo approaches, we use $32\times 16$ resolution environment light following NeRFactor.
Also, we use equal-area stratified sampling over hemispheres with 64 samples.
% The results for MC + Env do not incorporate the albedo prior, as it achieves better performance.
% We report the results for MC + Env with $\mathcal{L}_{\text{prior}}$ in the Appendix.

We report the quantitative results of the novel-view synthesis in Table~\ref{tab:view_synthesis} and intrinsic decomposition in Table~\ref{tab:intrinsic_decomposition_only} in terms of MSE, PSNR, and SSIM.
IBL-NeRF models outgoing radiance as the combination of various intrinsic components and concurrently generates images whose quality is comparable to vanilla NeRF.
Notably, our approach outperforms the method from NeRFactor (MC + Env) in both intrinsic decomposition and image synthesis results for all error metrics, which supports our claim that using environment lighting with MC sampling is inadequate to express complex indoor scenes.
The reconstruction quality is much better by alleviating the environment light and instead adapting our formulation in Eq.~\ref{eq:approximated_rendering_eq}.
Theoretically, the MC baseline should have better results in the expense of computation time, which is almost 3 times slower in the training phase and 5 times slower in the inference phase than IBL-NeRF.
However, since there exists a number of invalid samples in the incident radiance that are invisible from training viewpoints, the decomposition of MC is comparable to ours.
The results for MC + Env do not incorporate the albedo prior, as it achieves better performance.
% We report the second baseline method (MC + Env) with $\mathcal{L}_{\text{prior}}$ in Appendix.
We report the second baseline method (MC + Env) with $\mathcal{L}_{\text{prior}}$ in supplementary material.

We demonstrate the qualitative results of novel-view synthesis and intrinsic decomposition in synthetic scenes in Fig.~\ref{fig:view_synthesis}, ~\ref{fig:intrinsic_decomposition_kitchen} and~\ref{fig:intrinsic_decomposition_livingroom}, real scenes in Fig.~\ref{fig:intrinsic_decomposition_real}.
Our approach and MC approach with prefiltered radiance field reconstruct high-quality images in novel viewpoints, which are comparable to vanilla NeRF.
On the other hand, objects in large-scale indoor scenes are often occluded by other structures within the scene, and therefore cannot be illuminated appropriately with environment light (MC + Env).
The quality of images is significantly worse as it suffers from notable dark and noisy artifacts created from missing viewpoints or ambiguous regions.
Fig.~\ref{fig:intrinsic_decomposition_real}, ~\ref{fig:intrinsic_decomposition_kitchen} and~\ref{fig:intrinsic_decomposition_livingroom} show that IBL-NeRF successfully decomposes the scene attributes in both synthetic and real-world scenes.
IBL-NeRF estimates low roughness at metallic surfaces, for example, the ventilator, metallic wall, knobs in the oven, and pots in \textsc{Kitchen} in Fig.~\ref{fig:intrinsic_decomposition_kitchen}, TV in \textsc{Livingroom2} in Fig.~\ref{fig:intrinsic_decomposition_livingroom}.
However, our method fails to discover metallic surface that does not have specular variation with respect to viewing direction in the training set. (For example, the fireplace in \textsc{Livingroom2} has consistent color in the training images.)
% Additional results are available in the supplementary material.

\begin{figure}
    \centering
    \includegraphics[width=0.95\linewidth]{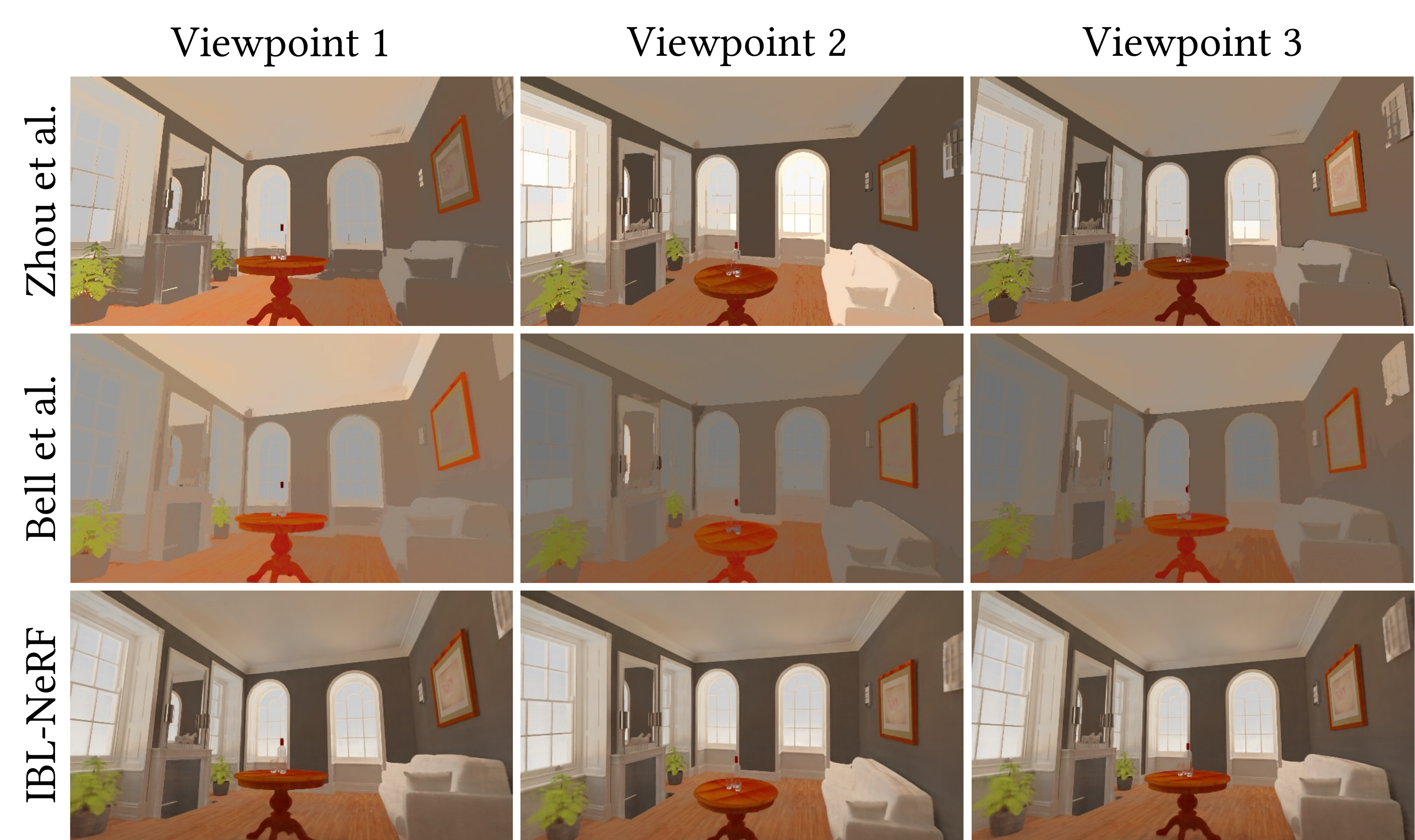}
    \caption{Visual comparison of albedo estimation between IBL-NeRF and single-image based methods.}
    \label{fig:albedo_consistency}
\end{figure}

Furthermore, IBL-NeRF can easily achieve the inherent multi-view consistency and smoothness of our optimizing process as shown in Fig.~\ref{fig:albedo_consistency}.
While the intrinsic decomposition algorithms for single-view images~\cite{bell2014intrinsic, zhou2015learning} fail to maintain consistent results, it provides useful guidance for the intrinsic decomposition.

\noindent\textbf{Ablation Studies}
Fig.~\ref{fig:intrinsic_decomposition_kitchen},~\ref{fig:intrinsic_decomposition_livingroom} and Table~\ref{tab:intrinsic_decomposition_only} also contain results for ablated versions of IBL-NeRF to analyze the important components of the proposed method.
The qualitative results with ground-truth normal $\mathbf{n}$ show cleaner roughness than our original model.
The effect of roughness is tightly coupled with the direction of mirror reflection, which is obtained from the surface normal.
Recent methods~\cite{Oechsle_2021_ICCV, NEURIPS2021_e41e164f} propose to reconstruct high-quality geometry with NeRF formulations, from which IBL-NeRF can learn better decomposition.

Since intrinsic decomposition is an under-constrained problem, prior knowledge on intrinsic components plays a crucial role to disambiguate each component.
When we remove $\mathcal{L}_{\text{prior}}$ the albedo contains illumination information which should belong to irradiance, and the irradiance is clipped to the mean value by $\mathcal{L}_{I,\text{reg}}$.
Also, without $\mathcal{L}_{I,\text{reg}}$, one cannot estimate correct irradiance, especially on the surface with dark albedo. (e.g., Oven in Fig.~\ref{fig:intrinsic_decomposition_kitchen} should have irradiance similar to nearby furniture, but the dark pixels encourage estimating lower irradiance without $\mathcal{L}_{I,\text{reg}}$.)
Removing both priors shows the worst results. 
% Without the prior loss, the accuracy of the view synthesis increases since the neural network can focus on minimizing the rendering loss with more flexibility.

\subsection{Scene Editing}

\begin{figure}
    \centering
    \subfloat[Original Scene]{\includegraphics[width=0.48\linewidth]{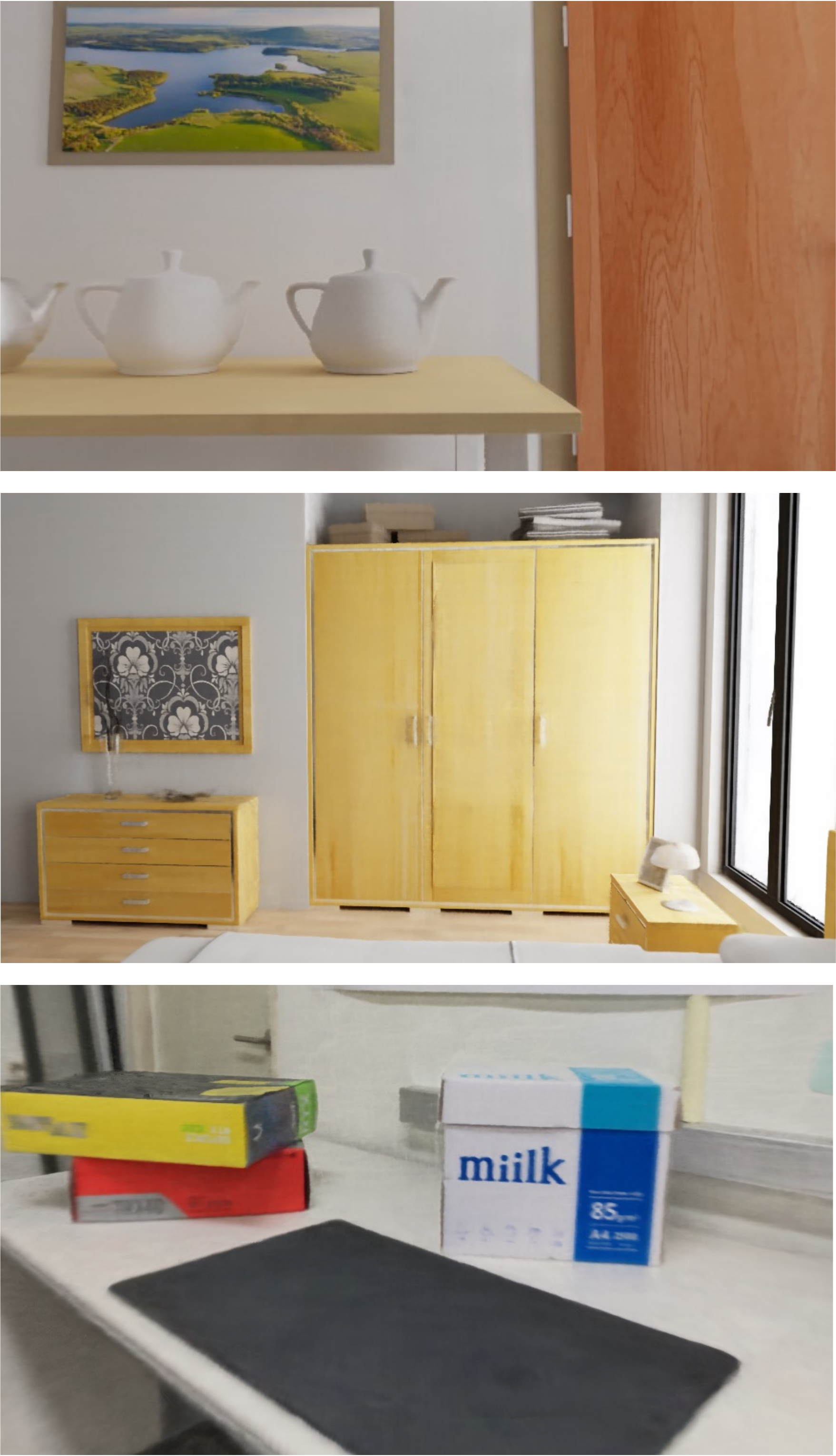}}\,
    \subfloat[Edited Scene]{\includegraphics[width=0.48\linewidth]{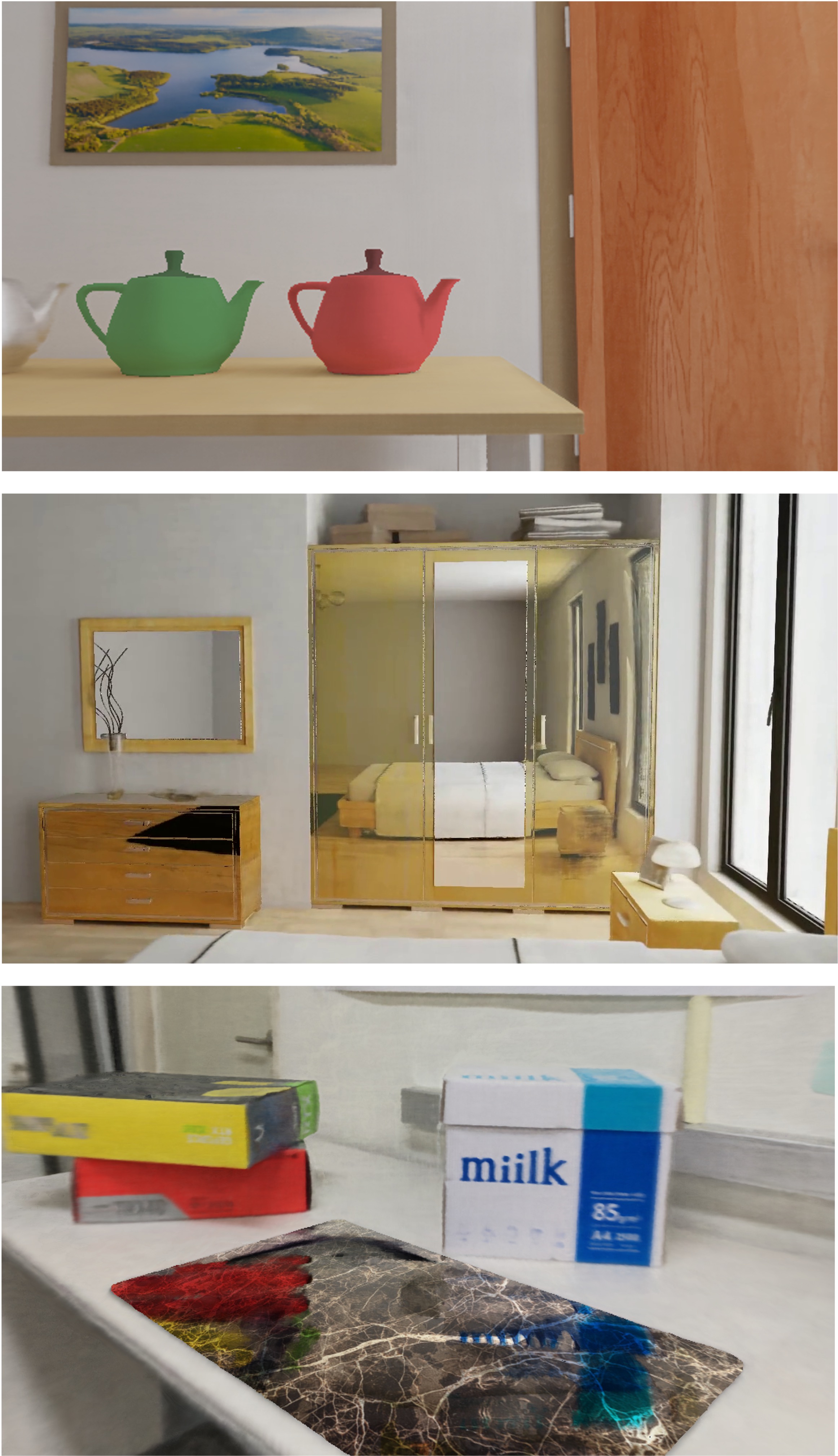}}
     \caption{Example of changing intrinsic components of the scene.}
     \label{fig:scene_edit}
\end{figure}

After IBL-NeRF decomposes intrinsic components, one can render realistic novel-view images of altered scenes by modifying the value of each component.
For example, we replace roughness of the dining table and albedo of the lamp to edit \textsc{Kitchen} scene in Fig.~\ref{fig:teaser}(d).
We demonstrate more results in Fig.~\ref{fig:scene_edit}.
In the first row of Fig.~\ref{fig:scene_edit},
we replace albedo of two kettles in \textsc{Veach-Ajar} to green and red respectively while preserving illumination information.
In the second row of Fig.~\ref{fig:scene_edit},
we reduce the roughness of the picture in frame, drawer, and closet door, which results in mirror-like material in \textsc{Bedroom} scene.
We also change the albedo of the middle door of the closet to white and the conference logo is marked on the left door by modifying roughness.
In the third row of Fig.~\ref{fig:scene_edit}, we modify the albedo and roughness of the desk pad in our real-world scene to express the marble-like material.
Also, one can insert 3D objects inside our trained neural volume with prefiltered radiance field.
In Fig.~\ref{fig:scene_add_obj}, we add 3 objects with different roughness and transparency inside the \textsc{Kitchen}.
The red blobby object is highly reflective and the surrounding scene is clearly reflected on its surface.
The green dragon also has a low roughness value but has translucency so the shape of the green kettle behind the object is visible.
Finally, the blue teapot has a high roughness value and moderate translucency.
The blurry reflection on the teapot accounts for its high roughness value.
Note that scene editing could be achieved similarly using Monte Carlo method with NeRF's radiance, but IBL-NeRF outperforms them in terms of speed (Table~{\ref{tab:view_synthesis}, Infer time).

\begin{figure}
    \centering
    \includegraphics[width=\linewidth]{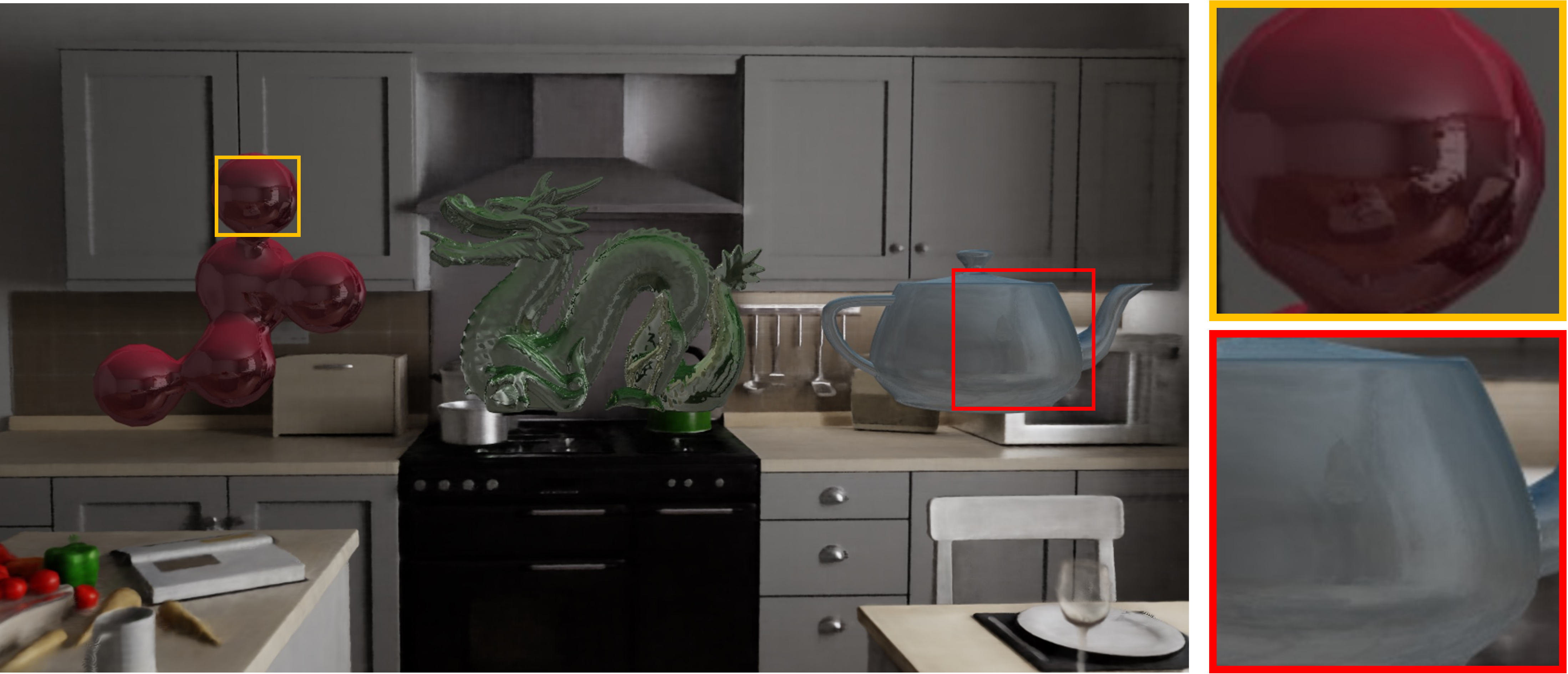}
    \caption{Example of adding new objects to the scene.}
    \label{fig:scene_add_obj}
    \vspace{-1em}
\end{figure}

%% file: tables/view_synthesis.tex
\begin{table}
\centering
\resizebox{\linewidth}{!}{
\begin{tabular}{l ccc cc}
\toprule
\multirow{2}{*}{Method} &
\multirow{2}{*}{MSE $\downarrow$} &
\multirow{2}{*}{PSNR $\uparrow$} &
\multirow{2}{*}{SSIM $\uparrow$} &
      \multicolumn{2}{c}{Time per step (s)} \\
      & & & & Train & Infer \\
\midrule
MC + Env~\cite{zhang2021nerfactor}	&	0.0369	&	16.107	&	0.2763 &	0.4686 &	0.1062\\
MC	&	0.0016	&	\textbf{30.052}	&	0.8348 &	0.4941 &	0.1084\\
NeRF &	\textbf{0.0008}	&	\textbf{34.707}	&	\textbf{0.9253} & \textbf{0.0984} & \textbf{0.0055}\\
%\midrule
IBL-NeRF 	&   \textbf{0.0014}	&	29.962	&	\textbf{0.9009}  &   \textbf{0.1559}  &   \textbf{0.0211}\\
\bottomrule
\end{tabular}
}
\caption{Quantitative results of view synthesis.}
\label{tab:view_synthesis}
\end{table}

% \begin{table}
% \centering
% \resizebox{0.95\linewidth}{!}{
% \begin{tabular}{@{}l ccc c@{}}
% \toprule
%     & MSE $\downarrow$   & PSNR $\uparrow$   & SSIM $\uparrow$   &  Time per step (s)\\ \midrule
% MC + Env (NeRFactor)	&	0.0369	&	16.107	&	0.2763 &	0.4686\\
% MC	&	0.0016	&	\textbf{30.052}	&	0.8348 &	0.4941\\
% NeRF &	\textbf{0.0008}	&	\textbf{34.707}	&	\textbf{0.9253} & \textbf{0.0984}\\
% \midrule
% IBL-NeRF 	&   \textbf{0.0014}	&	29.962	&	\textbf{0.9009}  &   \textbf{0.1559}\\
% \bottomrule
% \end{tabular}
% }
% \caption{Quantitative results of view synthesis.}
% \label{tab:view_synthesis}
% \end{table}

%% file: tables/intrinsic_decomposition_only.tex
\begin{table*}
\centering
\resizebox{0.85\linewidth}{!}{
\begin{tabular}{l|ccc|ccc|ccc}
\toprule
    & \multicolumn{3}{c|}{Albedo} & \multicolumn{3}{c|}{Irradiance} & \multicolumn{3}{c}{Roughness}\\ \cmidrule{2-10}
    & MSE $\downarrow$   & PSNR $\uparrow$   & SSIM $\uparrow$   & MSE $\downarrow$   & PSNR $\uparrow$   & SSIM $\uparrow$   & MSE $\downarrow$   & PSNR $\uparrow$   & SSIM $\uparrow$   \\ \midrule
MC + Env (NeRFactor)	&	0.1808	&	8.1273	&	0.3916	&	0.1190	&	10.220	&	0.2514	&	0.0910	&	11.798	&	0.6217\\
MC	&	\textbf{0.0543}	&	14.109	&	0.7383	&	\textbf{0.0344}	&	\textbf{17.280}	&	0.7149	&	0.0722	&	14.090	&	0.7474\\
IBL-NeRF 	&   0.0553	&	\textbf{14.114}	&	\textbf{0.7455}  &   0.0351	&	16.435	&	\textbf{0.7778}  &   \textbf{0.0707}	&	\textbf{15.545}	&	\textbf{0.8653}\\
\midrule
w/ GT $\mathbf{n}$	&	0.0551	&	\textbf{14.134}	&	\textbf{0.7465}  &   0.0376	&	15.986	&	0.7717  &   \textbf{0.0623}	&	14.216	&	0.8220\\
w/o $L_{\text{prior}}$ &	0.0664	&	13.423	&	0.7107	&	0.0403	&	15.609	&	0.7553	&	0.0717	&	15.413	&	0.8613\\
w/o $L_{I,\text{reg}}$ &   0.0551	&	14.077	&	0.7362	&	\textbf{0.0337}	&	16.215	&	0.7586	&	0.0710	&	14.316	&	0.8588\\
w/o all priors &   0.0775	&	11.601	&	0.6911	&	0.0674	&	12.147	&	0.7015	&	0.0709	&	15.527	&	0.8637\\
\bottomrule
\end{tabular}
}
\caption{Quantitative results of intrinsic decomposition.
% \jkim{(r1) more explanation?}
Compared to MC + Env (recent works on intrinsic decomposition), our result achieves better intrinsic decomposition of the scene. Note that MC shows comparable performance in the expense of substantially longer training time as it performs Monte-Carlo integration (Table~\ref{tab:view_synthesis}). Normal information is critical to estimate the shading information (irradiance, roughness), and the additional regularization on albedo or irradiance balances the distribution between different intrinsic components.}
\label{tab:intrinsic_decomposition_only}
\end{table*}

%% file: 05_conclusion.tex
\section{Conclusion}
We propose IBL-NeRF, a neural volume representation with prefiltered radiance field.
Our approach successfully decomposes the intrinsic components in a large-scale scene with an efficient approximation and prefiltered radiance field, which could not be processed in prior works with Monte Carlo integration of environment light.
Furthermore, one can easily edit the scene by modifying each decomposed component or inserting 3D models in our neural volume.
% Our illumination is baked within the neural volume as irradiance and prefiltered radiance field without explicit models of light sources.
% While this makes the approach efficiently capture and render complex global lighting with spatial variations, relighting is difficult.
Although IBL-NeRF can handle both Lambertian reflection and specular reflection, IBL-NeRF has a limitation in expressing transparent objects or perfect-mirror reflection.
One can resolve the ambiguity in a mirror with user interaction as~\cite{guo2021nerfren}.

%% file: 11_BRDF.tex
\section{BRDF Model}
\label{sec:BRDF}

IBL-NeRF adapts the microfacet BRDF model of Unreal Engine~\cite{karis2013real} and approximates the surface reflectance property with a set of decomposed intrinsic terms.
The BRDF (bidirectional reflectance distribution function) $f_r(\mathbf{x}, {\omega}_i, {\omega}_o)$ is the ratio between the incoming and outgoing radiance and it is a function of the surface location $\mathbf{x}$, incoming direction $\omega_i$, and the outgoing direction  $\omega_o$. 
Then the outgoing radiance is evaluated by attenuating the incoming radiance $L_i$ by the BRDF $f_r$, cosine term $\mathbf{n}\cdot \omega_i$ and integrating over all of the incoming directions~\cite{kajiya1986rendering}
\begin{equation}
    L_o(\mathbf{x}, \omega_o) = \int_{\Omega} f_r(\mathbf{x},\omega_i, \omega_o)L_i(\mathbf{x},\omega_i)(\mathbf{n}\cdot \omega_i)d\omega_i,
    \label{eq:radiance}
\end{equation}
where $\mathbf{n}$ is the surface normal at $\mathbf{x}$.
The radiance is composed of two terms, namely the diffuse term $L_{o,\text{diff}}$ and the specular term $L_{o,\text{spec}}$.
The distinction between the two terms stems from the conventional parameterization of BRDF into the two respective terms $f_r = f_{\text{diff}} + f_{\text{spec}}$.

\subsection{Diffuse Reflection}
\label{subsec:diffuse}

We use a simple Lambertian model for the diffuse BRDF
\begin{equation}
    f_{\text{diff}}  = (1-m) (1-F({\omega}_o, h))\frac{a}{\pi},
    \label{eq:diffuse_brdf}
\end{equation}
% where $a$ is the surface albedo and $m$ parameterizes how metallic it is\footnote{Metal does not refract at all, so $f_{\text{diff}}$ is attenuated by $(1-m)$.}.
where $a$ is the surface albedo and $m$ parameterizes how metallic it is.
While $a$ and $m$ are constants over incident angles, the diffuse term also varies according to  viewing directions due to the Fresnel effect $F$. 
The Fresnel effect models how much light is reflected and refracted.
% On the other hand, the diffuse term considers the directional components from Fresnel effect $F$ that determines how many light get reflected and refracted.
% $h = \frac{{\omega}_o + {\omega}_i}{\|{\omega}_o + {\omega}_i\|}$ is the halfway vector between $\omega_i$ and $\omega_o$.
% $F$ represents the Fresnel effect (cite?), and the Unreal Engine employs the Fresnel-Schlick approximation~\cite{schlick1994inexpensive} 
The Unreal Engine employs the Fresnel-Schlick approximation~\cite{schlick1994inexpensive}
\begin{equation} \label{eq:Fresnel_schlick}
    F({\omega}_o, \mathbf{h}) = F_0 + (1-F_0)(1-({\omega}_o \cdot \mathbf{h}))^5,
\end{equation}
where $\mathbf{h} = \frac{{\omega}_o + {\omega}_i}{\|{\omega}_o + {\omega}_i\|}$ is the halfway vector and $F_0= \text{lerp}(0.04, a, m)$ is a constant term defined as the linear interpolation between $0.04$ and $a$.

Now we can find the diffuse component of the outgoing radiance by combining Eq.~(\ref{eq:radiance}) with (\ref{eq:diffuse_brdf}):
\begin{multline}
    L_{o, \text{diff}}(\mathbf{x}, {\omega}_o) \\
    = (1-m)\frac{a}{\pi} \int_\Omega (1-F({\omega}_o, \mathbf{h}))  L_i(\mathbf{x},\omega_i)(\mathbf{n}\cdot \omega_i) d\omega_i,% &\simeq (1-m) (1-F(\omega_o,n))\frac{a}{\pi} \int_\Omega L_i(x,\omega_i)(n\cdot \omega_i) d\omega_i.
\end{multline}
since $a$ and $m$ are constants over $\omega_i$.
We can further simplify the integrand by assuming a constant Fresnel term.
One na\"ive way is to replace the dependency of $\mathbf{h}$ with $\mathbf{n}$, and substitute $F(\omega_o,\mathbf{n})$.
But such formulation largely deviates from the diffuse property and incurs excessive reflection near the edge.
% To alleviate the phenomenon, we follow common techniques in many commercial rendering engines and inject roughness $\gamma$ to the Fresnel term (cite)
To alleviate the phenomenon, we inject roughness $\gamma$ to the Fresnel term as  ~\cite{lagarde2011}
\begin{equation}
    F_{\gamma}(\omega_o,\mathbf{n},\gamma) = F_0 + (\max(1-\gamma, F_0) - F_0) (1-(\mathbf{n} \cdot \omega_o))^5.
    \label{eq:fresnel_approx}
\end{equation}
After factoring out the approximated Fresnel term $F_\gamma$, our diffuse component is formulated as following:
\begin{equation}
     (1-m)(1-F_{\gamma}(\omega_o,\mathbf{n},\gamma))a  \underbrace{ \left\{\frac{1}{\pi}\int_\Omega L_i(\mathbf{x},\omega_i)(\mathbf{n}\cdot \omega_i) d\omega_i \right\} }_{\text{Irradiance }I}.
\end{equation}
The remaining integrand is the sum of total incident radiance received at point $\mathbf{x}$, which is also known as the \textit{irradiance} $I$.
% In real-time image based rendering, irradiance is precalculated and stored as an additional environment map, fetched at runtime using normal direction $n$.
In real-time image-based rendering, irradiance is fetched at the normal direction $\mathbf{n}$ from an additional environment map, which is precalculated and stored.
However, IBL-NeRF implicitly represents irradiance at each position using MLP instead of an environment map.

\subsection{Specular Reflection}
\label{subsec:specular}

The specular BRDF follows the Cook-Torrance model~\cite{cook1982reflectance}
\begin{equation}
    f_{\text{spec}}  = \frac{D(\mathbf{h},\mathbf{n},\gamma) \cdot F({\omega}_o, \mathbf{h}) \cdot G({\omega}_i, {\omega}_o, \mathbf{n}, \gamma)}{4(\mathbf{n} \cdot {\omega}_o)(\mathbf{n} \cdot {\omega}_i)}.
    \label{eq:specular_brdf}
\end{equation}
Note that the roughness term $\gamma$, introduced in the approximate Fresnel term in Eq.~(\ref{eq:fresnel_approx}), plays a crucial role in the specular reflection.
In addition to the Fresnel equation defined in Eq.~(\ref{eq:Fresnel_schlick}), the specular reflection is also dependent on the normal distribution function $D$ and the geometry function $G$. %, which are also used in Unreal Engine.
The normal distribution is adopted from Trowbridge-Reitz GGX~\cite{trowbridge1975average}
\begin{equation}
    D(\mathbf{h},\mathbf{n},\gamma) = \frac{\alpha ^2}{\pi((\mathbf{n} \cdot \mathbf{h})^2(\alpha^2 - 1)+1)^2},
    \label{eq:normal_distribution}
\end{equation}
where  $\alpha = {\gamma}^2$. The geometry function $G$ describes self-shadowing according to Smith's Schlick-GGX~\cite{smith1967geometrical}
\begin{equation}
    G({\omega}_i, {\omega}_o, \mathbf{n}, \gamma) = G_{\text{Schlick}}({\omega}_i, \mathbf{n}, \gamma) G_{\text{Schlick}}({\omega}_o, \mathbf{n}, \gamma), 
\end{equation}
where $G_{\text{Schlick}}({\omega}, \mathbf{n}, \gamma) = \frac{(\mathbf{n}\cdot \omega)}{(\mathbf{n}\cdot \omega)(1-k)+k}$ and $k=\frac{\alpha^2}{2}$.
% \begin{equation*}
%     \mbox{where}\quad G_{\text{Schlick}}({\omega}, n, \gamma) = \frac{(n\cdot \omega)}{(n\cdot \omega)(1-k)+k}\quad \mbox{and} \quad k= \frac{\alpha^2}{2}.
% \end{equation*}
Basically, roughness $\gamma$ affects the sharpness of angular distribution of reflected radiance.

\newcommand{\pdf}{p(\omega_i |\mathbf{x}, \mathbf{n}, \omega_o, \gamma)}
\newcommand{\pdfk}{p(\omega_i^k |\mathbf{x}, \mathbf{n}, \omega_o, \gamma)}
\newcommand{\wipdf}{\omega_i \sim \pdf}

The outgoing radiance of specular component involves a highly complex distribution compared to the diffuse component.
The specular term can be computed using importance sampling as follows:
\begin{equation*}
    L_{o, \text{spec}}(\mathbf{x}, {\omega}_o) = \frac{1}{N} \sum_{k=1}^{N} \frac{f_{\text{spec}}(\mathbf{x}, \omega_i^k, \omega_o) L_i(\mathbf{x},\omega_i)(\mathbf{n}\cdot \omega_i^k)}{\pdfk}.
\end{equation*}
% \begin{align*}
%     L_{o, \text{spec}}(x, {\omega}_o) &= \int_\Omega f_{\text{spec}}(x, \omega_i, \omega_o) L_i(x,\omega_i)(n\cdot \omega_i) d\omega_i \\
%     &= \frac{1}{N} \sum_{k=1}^{N} \frac{f_{\text{spec}}(x, \omega_i^k, \omega_o) L_i(x,\omega_i)(n\cdot \omega_i^k)}{\pdfk}.
% \end{align*}
% \begin{align*}
%     & L_{o, specular}(p, {\omega}_o) \\ &= \int_\Omega f_{spec}(p, \omega_i, \omega_o) L_i(p,\omega_i)(n\cdot \omega_i) d\omega_i \\
%     &= {\mathbb{E}}_{\omega_i\sim p(\omega_i|p,\omega_o)}
%     \left[ \frac{f_{spec}(p, \omega_i^k, \omega_o) L_i(p,\omega_i)(n\cdot \omega_i^k)}{p(\omega_i^k|p,\omega_o)} \right] \\
%     &= {\mathbb{E}}_{\omega_i\sim p(\omega_i|p,\omega_o)}
%     \left[ L_i(p,\omega_i) \right]
%     {\mathbb{E}}_{\omega_i\sim p(\omega_i|p,\omega_o)}
%     \left[ \frac{f_{spec}(p, \omega_i^k, \omega_o) (n\cdot \omega_i^k)}{p(\omega_i^k|p,\omega_o)} \right]
% \end{align*}
The sampling PDF $\pdf$ can be found from the normal distribution function $D$. ~\cite{walter2007microfacet}. %$D(h, n,\gamma) (h \cdot n)$ in Eq.~(\ref{eq:normal_distribution})
We split the above sum into two components, adapting the \textit{split-sum approximation} ~\cite{karis2013real}
\begin{equation}
    \underbrace{ \left\{ 
    \frac{1}{N} \sum_{k=1}^{N} L_i(\mathbf{x},\omega_i^k)
    \right\}}_{\simeq L_\mathrm{pref}(\mathbf{x},\omega_r, \gamma)}
   \underbrace{ \left\{ 
   \frac{1}{N} \sum_{k=1}^{N} \frac{f_{\text{spec}}(\mathbf{x}, \omega_i^k, \omega_o)(\mathbf{n}\cdot \omega_i^k)}{\pdfk} 
    \right\}}_{\simeq [F_0,1]^T \mathrm{LUT}(\omega_o, \cdot \mathbf{n}, \gamma)}.
    \label{eq:specular_approx}
\end{equation}
The approximation separates the effect of lighting (first term) from BRDF (second term), and allows accelerated computation of radiance  with precalculated maps.
% We further discuss our efficient approximation of the two term in the following subsections.
We further discuss the approximation of the two terms in the following subsections.

\subsubsection{Prefiltered Radiance Fields}
% The first term is dependent on $\omega_o$, since the sampling PDF $\pdf$ is dependent on the viewing direction $\omega_o$.
% The dependency on the viewing direction greatly increases the amount of computation, while its effect is minimal except for the reflection from an extremely grazing angle.
% It is often assumed that $\omega_o=\omega_r=n$, where  $\omega_r=\text{mirror}(\omega_o, n)$ is the direction of mirror reflection of $\omega_o$ with respect to the surface normal.
% %To remove the dependency on the viewing  direction, it is often assumed that $n=\omega_o=\omega_r$ where $\omega_r=\text{mirror}(\omega_o, n)$.
% % This approximation makes quality of reflections at grazing angle worse, but greatly reduces complexity.
% With this approximation, now the first term only depends on the roughness value $\gamma$.
% We pre-calculate the first term for different roughness values and store it as an environment map with several mipmap levels, which is known as \textit{prefiltered environment map} $L_{\text{pref}}$ \cite{karis2013real}.
% % Note that if roughness becomes $1$, $L_{\text{pref}}$ is close to irradiance.
% Similar to texture mipmap, we can fetch the prefiltered map of the desired roughness in the mirror-reflected direction $\omega_r$ using trilinear interpolation as
% \begin{equation}
%     L_{\text{pref}}(x, \omega_r, \gamma) = \sum_i w^i(\gamma) L^i_{\text{pref}}(x, \omega_r), 
% \end{equation}
% where $L^i_{\text{pref}}$ is $i$th mipmap and $w^i(\gamma)$ is weight of $\gamma$ to $i$th mipmap.
Considering the sampling PDF, the first term is dependent on three terms $\left[\omega_o, (\omega_o \cdot \mathbf{n}), \gamma \right]$.
Such multiple dependencies greatly increase the amount of computation and memory required to store precalculated results.
%, while its effect is minimal except for the reflection from an extremely grazing angle.
Thus, it is often assumed that $\omega_o=\mathbf{n}=\omega_r$, where  $\omega_r=\text{mirror}(\omega_o, \mathbf{n})$ is the direction of mirror reflection of $\omega_o$ with respect to the surface normal. %\footnote
{Note that $\omega_r$ is chosen since $\pdf=\delta (\omega_r)$ if $\gamma=0$.}
%($\pdf=p(\omega_i|x, \omega_r, \gamma)$).
%To remove the dependency on the viewing  direction, it is often assumed that $n=\omega_o=\omega_r$ where $\omega_r=\text{mirror}(\omega_o, n)$.
% This approximation makes quality of reflections at grazing angle worse, but greatly reduces complexity.
With this isotropic approximation, now the first term only depends on $\omega_r$ and $\gamma$.
We precalculate the first term for each direction $\omega_r$ with different roughness values and store it as an environment map with several mipmap levels, which is known as \textit{prefiltered environment map} $L_{\text{pref}}$ \cite{karis2013real} \footnote{We will consider $L_{\text{pref}}$ from incident direction similar to $L_i$}.
% Note that if roughness becomes $1$, $L_{\text{pref}}$ is close to irradiance.
Similar to texture mipmap, we can fetch the prefiltered radiance of the desired roughness in the mirror-reflected direction $\omega_r$ using trilinear interpolation as
\begin{equation}
    L_{\text{pref}}^j(\mathbf{x}, \omega_r, \gamma) = \sum_j w^j(\gamma) L^j_{\text{pref}}(\mathbf{x}, \omega_r), 
\end{equation}
where $L^j_{\text{pref}}$ is $j$th mipmap and $w^j(\gamma)$ is the weight of $\gamma$ for $j$th mipmap.
In IBL-NeRF, the prefiltered environment map is stored implicitly in MLP, as irradiance in the previous section.
Therefore we will rather refer it as \textit{prefiltered radiance fields}.

\subsubsection{Precomputing BRDF Integration}
By substituting the Fresnel term in Eq.~(\ref{eq:Fresnel_schlick}), the second term in Eq.~(\ref{eq:specular_approx}) can be formulated as an affine function of $F_0$:
\begin{align*}
   \int_\Omega  & f_{\text{spec}}  (\mathbf{x}, \omega_i, \omega_o) (\mathbf{n}\cdot \omega_i) d\omega_i \\ &= F_0 \int_\Omega \frac{f_{\text{spec}}(\mathbf{x}, \omega_i, \omega_o)}{F(\omega_o,\mathbf{h})} (1-(1-(\omega_o \cdot \mathbf{h}))^5) (\mathbf{n}\cdot \omega_i) d\omega_i \\ & \quad \quad + \int_\Omega \frac{f_{\text{spec}}(\mathbf{x}, \omega_i, \omega_o)}{F(\omega_o,\mathbf{h})} (1-(\omega_o \cdot \mathbf{h}))^5 (\mathbf{n}\cdot \omega_i) d\omega_i \\
   & = [F_0, 1]^T \mathrm{LUT} (\omega_o \cdot \mathbf{n}, \gamma).
\end{align*}
After integration over $\Omega$, we can remove the dependency on $\omega_i$, and the scale and bias term depend only on $\gamma$ and $(\mathbf{n} \cdot \omega_o)$~\cite{karis2013real}.
The integral can be precalculated and stored as a 2D lookup texture (LUT), as illustrated in Fig.2 (g) in main manuscript.

% To sum up, specular component could be approximated as follows.
% \begin{equation}
%     L_{o, specular}(p, {\omega}_o) = L_{pref}(p, \omega_r, \gamma) \times [F_0, 1]^T LUT(\omega_o \cdot n, \gamma)
% \end{equation}

\subsection{Final Radiance Approximation}

Combining previous sections, the outgoing radiance $L_o(x, {\omega}_o)$ is approximated as following without Monte Carlo integration,
\begin{align} \label{equation:final_approximation}
    L_o(\mathbf{x}, {\omega}_o) &= (1-m)\times(1-F_{\gamma}(\omega_o,\mathbf{n},\gamma))\times a\times I \nonumber \\ 
    &+ L_{\text{pref}}(\mathbf{x}, \omega_r, \gamma) \times [F_0, 1]^T \text{LUT}(\omega_o \cdot \mathbf{n}, \gamma),
\end{align}
 where the first term is the diffuse component and the second term is the specular component.
 The formulation in Eq.~(\ref{equation:final_approximation}) is also visualized in Fig.2 of main manuscript.
 Given the precomputed maps, we only need to estimate the surface normal, albedo, irradiance, and roughness of the scene to find the diffuse and specular components.

%% file: 12_prefiltered_radiance.tex
\section{Prefiltered Radiance} % 본문에 못 적은 내용들
\label{appendixb}
\subsection{Prefiltered Radiance in Image Space}

\begin{figure}
    \centering
    \includegraphics[width=0.9\linewidth]{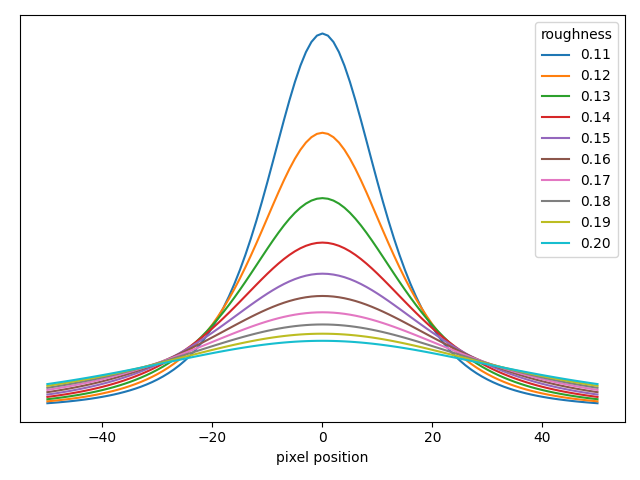}
    \caption{Kernel shapes in screen space for different roughness values.}
    \label{fig:kernel}
\end{figure}

In this section, we discuss the choice of Gaussian convolution kernels, which  approximates the radiance fields with different roughness $\gamma$.
The prefiltered radiance $L_\mathrm{pref}$ in our approximation is calculated in the image space $S$ of the current view.
Recall that, for specular reflection, the radiance field is fetched from the set of the radiance values with different sharpness, according to the roughness value of the surface point.
The radiance value for $\mathbf{x}$ along the direction $\omega$ is
\begin{align}
    L_{\text{pref}}^j(\mathbf{x},\omega) & = \mathbb{E}_{\omega_i\sim p(\omega_i|\mathbf{x}, \omega, \gamma_j)} [L_i(x, \omega_i)] \\
    & = \int_\Omega L_i(\mathbf{x}, \omega_i) p(\omega_i|\mathbf{x}, \omega, \gamma_j) d \omega_i \\
    & = \int_S L_i(s_i) p_S(s_i|\mathbf{x}, \omega, \gamma_j) d s 
\end{align}
where $s_i$ is the screen space coordinate that corresponds to direction $\omega_i$.
The $j$th radiance field contains the approximated values for the roughness value of $\gamma_j$.
The sampling probability on the screen space $p_S$ could be calculated as following
\begin{align}
    & p_S(s_i|\mathbf{x},\omega, \gamma_j) = p(h|\mathbf{x},\omega,\gamma_j) \norm{\frac{\partial h}{\partial \omega_i}} \norm{\frac{\partial \omega_i}{\partial s_i}} \\
    &= D(\mathbf{h},\omega,\gamma_j) (\mathbf{h} \cdot \omega) \left( \frac{1}{4(\omega_i \cdot \mathbf{h})} \right) \left( \frac{(\omega_i \cdot v)}{f^2+s_i^2} \right),
\end{align}
where $\mathbf{h}$ is the halfway vector between $\omega$ and $\omega_i$, $D$ is the normal distribution function, $v$ is the viewing direction of the camera, and $f$ is focal length of the camera.
$D$ is the filter kernel, and additional terms are Jacobians to map halfway vectors into pixel coordinates.
Now we assume $\omega=v$ in order to use a globally consistent convolution kernel.
Then the convolution kernel for roughness $\gamma_j$ in the image space can be designed as  
\begin{equation}
    K^j(s_i) \propto p_S(s_i|\mathbf{x},\omega,\gamma_j)
\end{equation}
for each pixel $s_i$.
Thus $L_{\text{pref}}^j(\mathbf{x},\omega)=K^j(L(s))$, where $s$ is the screen space coordinate that corresponds to $\omega$.
The examples of $K^j$ are plotted in Fig.~\ref{fig:kernel} for different roughness values.
The overall shape is similar to that of Gaussian function, which is used to approximate $L_{\text{pref}}^j(\mathbf{x},\omega)$ in our implementation.
We did not rigorously calibrate the parameters of Gaussian functions to $K^j$, which could be further studied in the future work.

\begin{figure}
    \centering
    \includegraphics[width=0.95\linewidth]{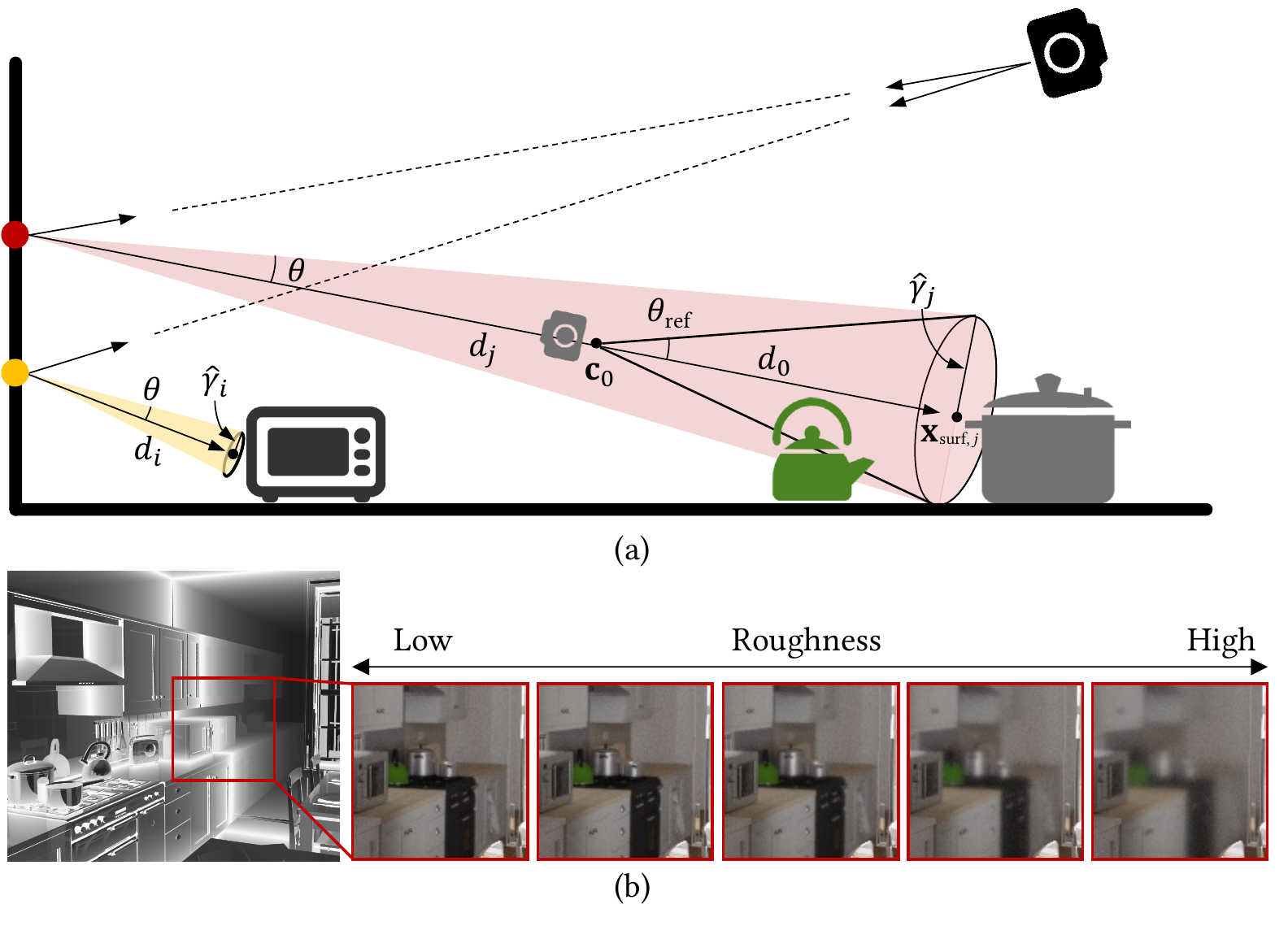}
    \caption{(a) The kernel sizes of image-space filtering depends on the observation angle ($\tan \theta$) while the effective kernel in the 3D spatial domain ($\hat{\gamma}$) depends on the distance to the object. 
    (b) We illustrate the specular reflection on the surface of the back wall in \textsc{Kitchen} scene.
    The left image shows the distance to the hit point reflected from the back wall, indicated with a red box.
    The right image shows the reflected radiance when the wall is assigned with different roughness values. Even though the roughness of the wall is constant, the reflection is blurred with different roughness $\gamma^\text{ref}$  that accounts for distance variations; the reflection of closer object  (microwave) is sharper, whereas the farther objects (kettle and pot) are blurrier.}
    \label{fig:prefiltered_explanation}
\end{figure}

\subsection{Convolution Level Adaptation}
While image-space filtering is an efficient means to aggregate neighboring rays, the approximate filter size to be applied on the image should depend on both the surface roughness and the distance that the reflected ray travels, denoted $d$.
The size of the image filter is determined by the tangent of the observation angle, $\tan \theta$ which is assumed to be proportional to the convolution level of prefiltered radiance $\gamma$.
Fig.~\ref{fig:prefiltered_explanation}(a) shows the effective kernel size for image-space filtering and the corresponding roughness values.
The reflection ray at the red point hits farther objects than the yellow point and needs to be blurred with a larger kernel even though observing the left wall having a constant roughness. 
When the filter is projected near the surface point $\mathbf{x}_\text{surf}$, the same observation angle $\theta$ corresponds to different sizes of effective range, denoted $\hat{\gamma} = d \tan \theta$, where $d$ is the distance to $\mathbf{x}_\text{surf}$.

However, it is infeasible to consider different $d$ values for every surface point shown on the radiance field, moreover, the distances are unknown before training the neural network.
Therefore we simply train $\L_{\text{pref}}$ assuming a constant distance $d_0$ for all the pixels, and we set $d_0$ to be the mean value of the near and far plane.
It is illustrated in Fig.~\ref{fig:prefiltered_explanation}(a) where the small virtual camera observes the hit point of reflected radiance from $d_0$, and therefore assuming the same $\hat{\gamma}$ with given level $\gamma$.
At the inference phase, we can calculate the reflected distance $d$ and compensate for the discrepancy compared to the trained depth $d_0$.
Specifically, we scale the inferred roughness $\gamma$ with the depth ratio and use $\gamma^\text{ref} = (d/d_0) \cdot \gamma$ to find the appropriate level of prefilter.
Fig.~\ref{fig:prefiltered_explanation}(b) shows the effect of depth on the normalized filter size. When the objects are farther from the hit point (kettle and pot), the reflected radiance is blurrier than the closer ones (microwave).

\begin{figure}
    \centering
    \includegraphics[width=\linewidth]{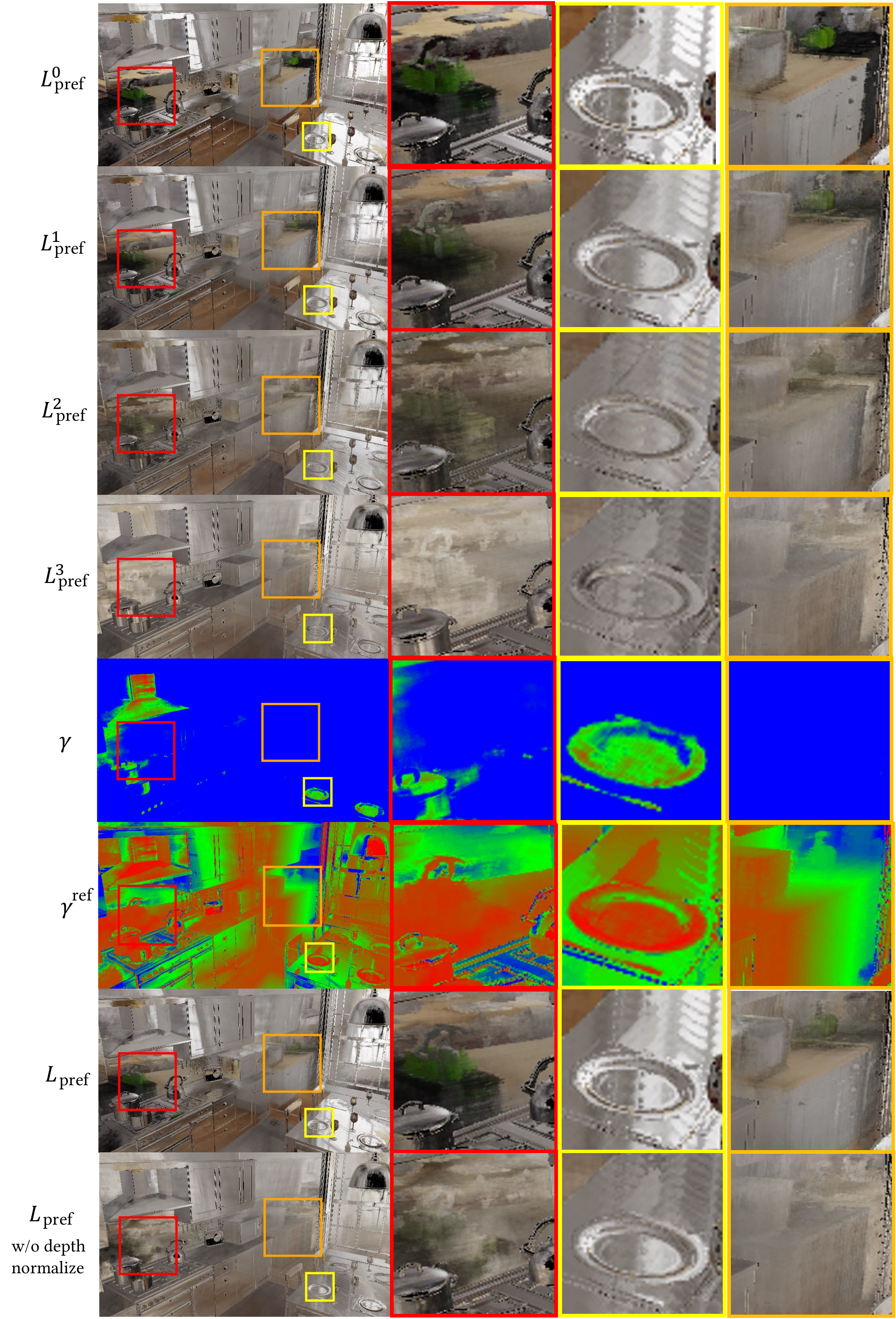}
    \caption{Prefiltered radiance with different levels ($L_\text{pref}^j$). The 5, 6th row shows the roughness value $\gamma$ and normalized roughness value $\gamma^{\text{ref}}$ respectively, where red indicates low and blue indicates high.
    The last two rows show the combined results after trilinear interpolation ($L_\text{pref}$) using $\gamma^{\text{ref}}$ and $\gamma$ respectively.}
    \label{fig:prefiltered}
\end{figure}

\subsection{Prefiltered Radiance at Reflected Direction}
Prefiltered radiance $L_{\text{pref}}$ at $\mathbf{x}_\text{surf}$ toward $\omega_r$ trained with screen-space prefiltered radiance approximation is visualized in Fig.~\ref{fig:prefiltered}.
Given the predefined set of roughness values $\gamma_j$, $L_\text{pref}^j$ represents the $j$th level prefiltered radiance.
First four rows of Fig.~\ref{fig:prefiltered} show $L_\text{pref}^j$ with different levels.
Since the higher level of $j$ performs convolution using a Gaussian kernel with a wider range, $L_\text{pref}^j$ with higher $j$ looks more blurry.
Fifth and sixth row show roughness $\gamma$ and normalized roughness ($\gamma^\text{ref} = (d/d_0) \cdot \gamma$) which are used for prefiltered radiance fetching index.
$d$ is distance from $\mathbf{x}_{\text{surf}}$ to the next hit point along $\omega_r$.
Last two rows show $L_\text{pref}$s, which is the result of trilinear interpolation of $L^j_{\text{pref}}$'s whose weights are deduced from $\gamma^\text{ref}$ and $\gamma$ respectively.
Note that without using normalization, the convolution occurs with a constant kernel regardless of $d$, which is not physically correct (orange box).

%% file: 13_additinal_results.tex
\section{Additional Results}

\begin{figure}
    \centering
    \includegraphics[width=\linewidth]{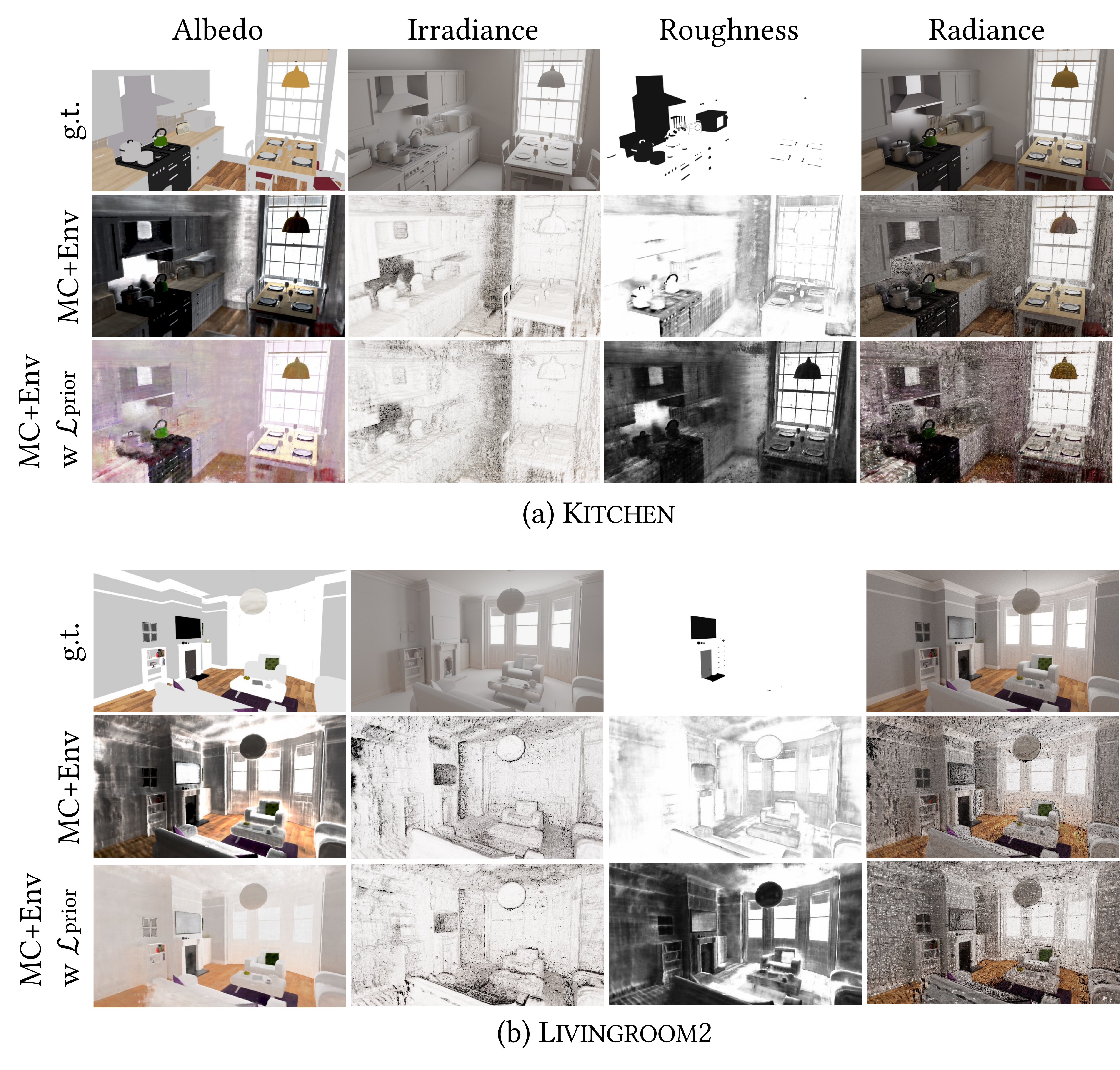}
    \caption{Qualitative results of MC + Env baseline with applying $\mathcal{L}_{\text{prior}}$. We observe that applying $\mathcal{L}_{\text{prior}}$ loss to MC + Env baseline worsens the reconstruction quality. Therefore, we do not apply $\mathcal{L}_{\text{prior}}$ to MC + Env baseline in the main experiments.}
    \label{fig:MC_env_prior_result}
\end{figure}
\subsection{Scene Editing}
\begin{figure*}
    \centering
    \includegraphics[width=\linewidth]{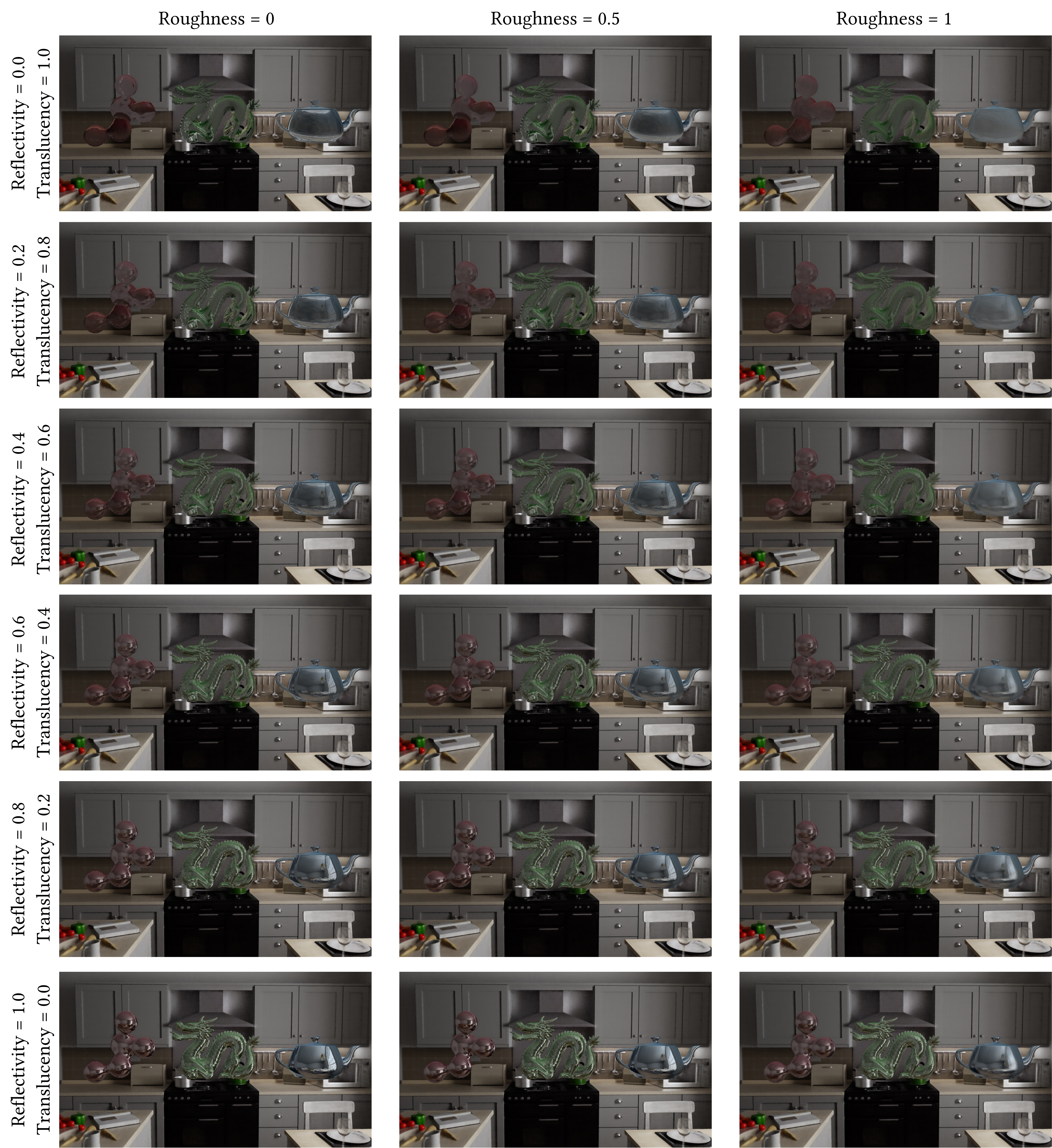}
    \caption{Additional samples of inserting new objects inside the scene.}
    \label{fig:scene_add_obj}
\end{figure*}

We display additional examples of inserting 3D object into our learned neural volume in Fig.~\ref{fig:scene_add_obj}. 
We vary roughness, reflectivity, and translucency with different levels.
Thanks to our prefiltered radiance, one can render high-quality image with different roughness and transparency.

\begin{figure}
    \centering
    \subfloat[Original Scene]{\includegraphics[width=0.47\linewidth]{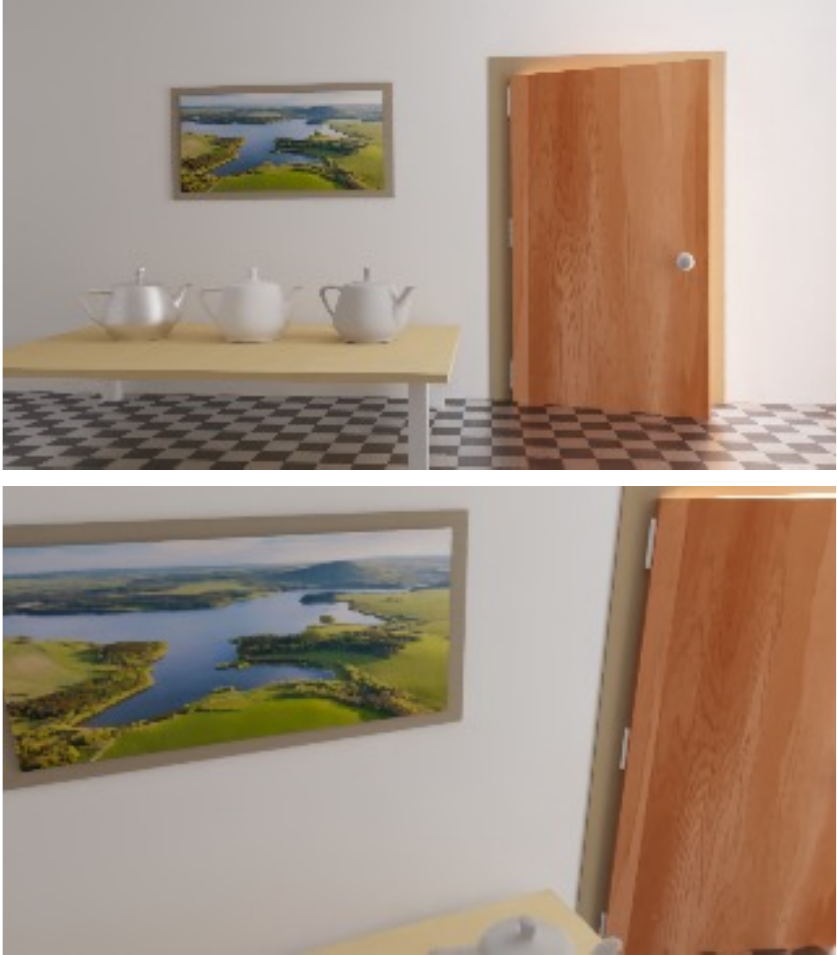}}\:
    \subfloat[Edited Scene]{\includegraphics[width=0.47\linewidth]{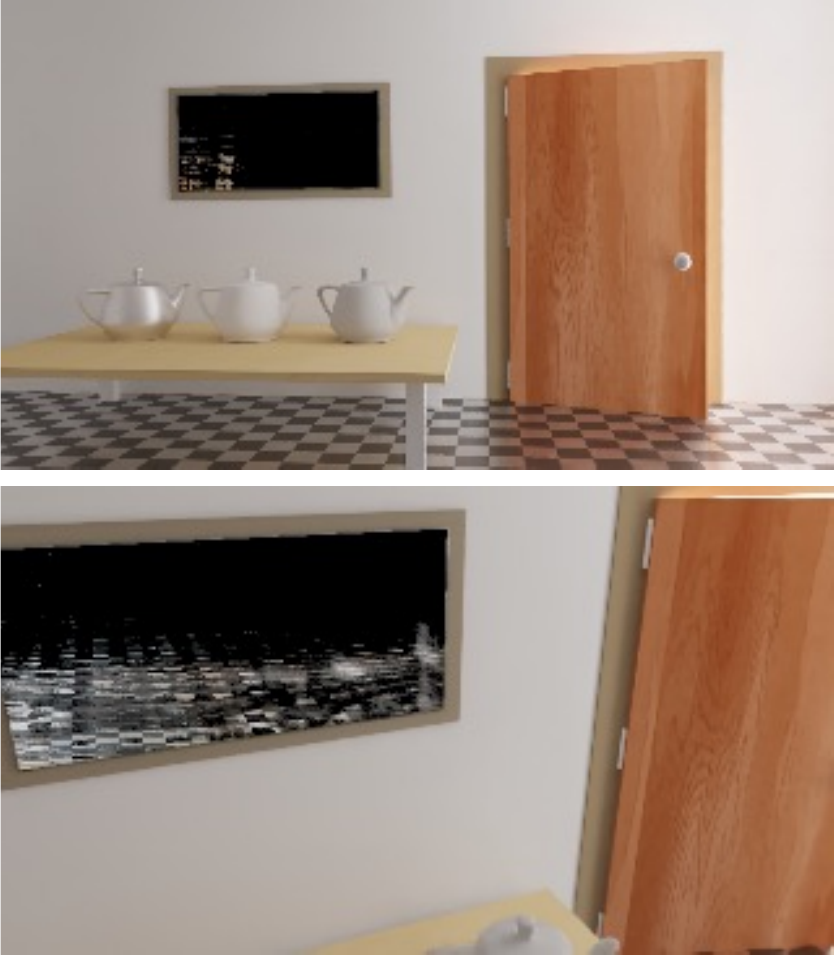}}
    \caption{Failure cases of our scene editing task.}
    \label{fig:scene_editing_failure}
\end{figure}

Also, we report a failure case of our scene editing task in Fig.~\ref{fig:scene_editing_failure}.
We lower the roughness of the picture in the frame in the \textsc{Veach-Ajar}.
The viewpoints in the training image set of the \textsc{Veach-Ajar} scene are highly restricted.
Most of the viewpoints are facing the front of the desk where the kettles are placed.
Since there is no visual information in the backward region, the prefiltered radiance is poorly optimized in the unseen area.
Therefore, failure in editing scenes with restricted viewpoint is a natural result.

\subsection{View Synthesis \& Intrinsic Decomposition}
% We add supplementary quantitative results of intrinsic decomposition for diffuse reflection ($L_{o,\text{diff}}$) and specular reflection ($L_{o,\text{spec}}$) in Table~\ref{tab:additional_intrinsic_decomposition}.
First, we report the second baseline method (MC + Env) with $\mathcal{L}_{\text{prior}}$ incorporated in Fig.~\ref{fig:MC_env_prior_result}.
We observe that Monte Carlo approach with environment light shows inferior intrinsic decomposition performance with our albedo prior loss.
Especially, MC + Env totally fails to estimate valid roughness with $\mathcal{L}_{\text{prior}}$.

\begin{figure*}
    \centering
    \includegraphics[width=\linewidth]{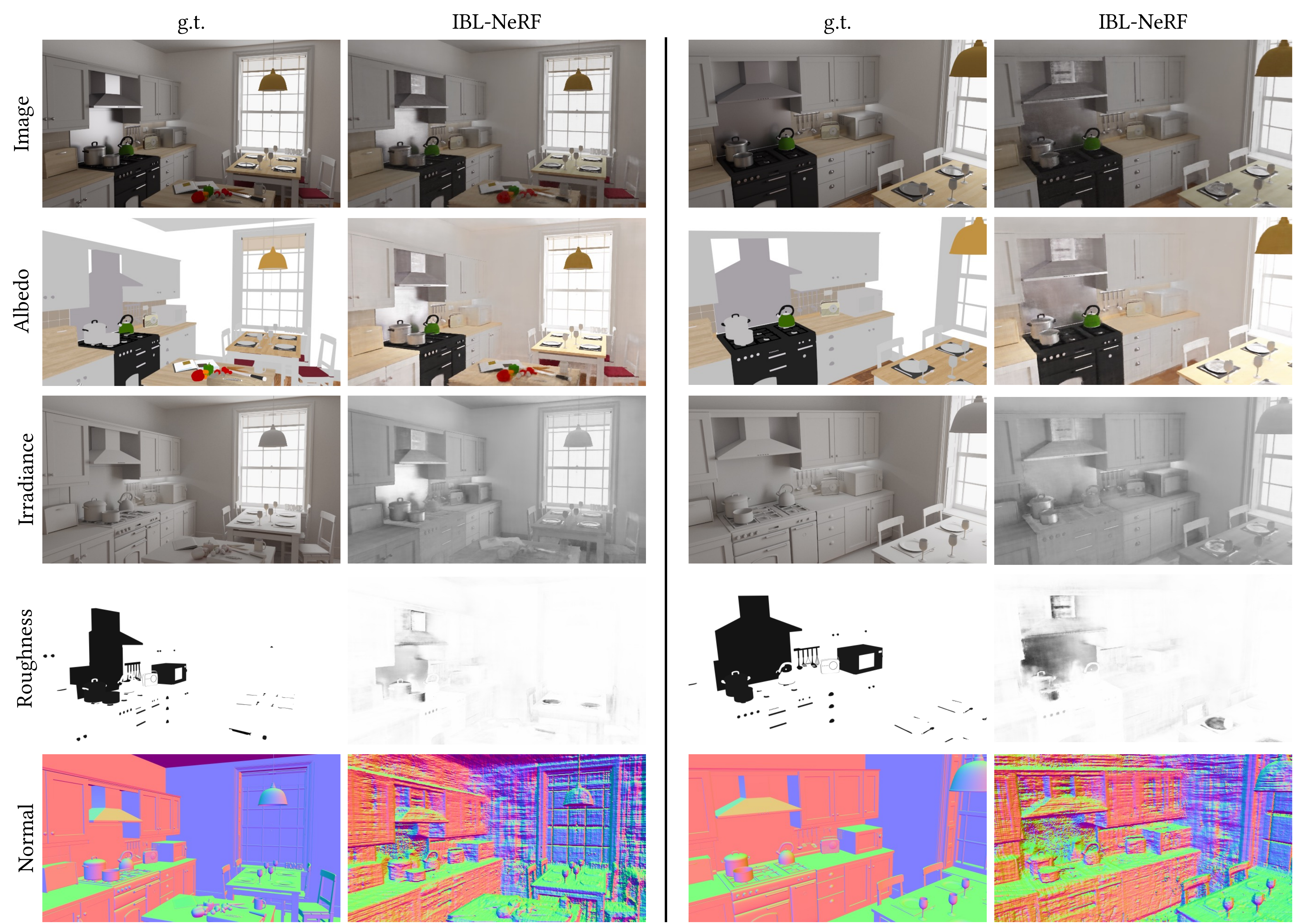}
    \caption{Additional qualitative results of novel view image synthesis and intrinsic decomposition in \textsc{Kitchen}.}
    \label{fig:additional_kitchen}
\end{figure*}

\begin{figure*}
    \centering
    \includegraphics[width=\linewidth]{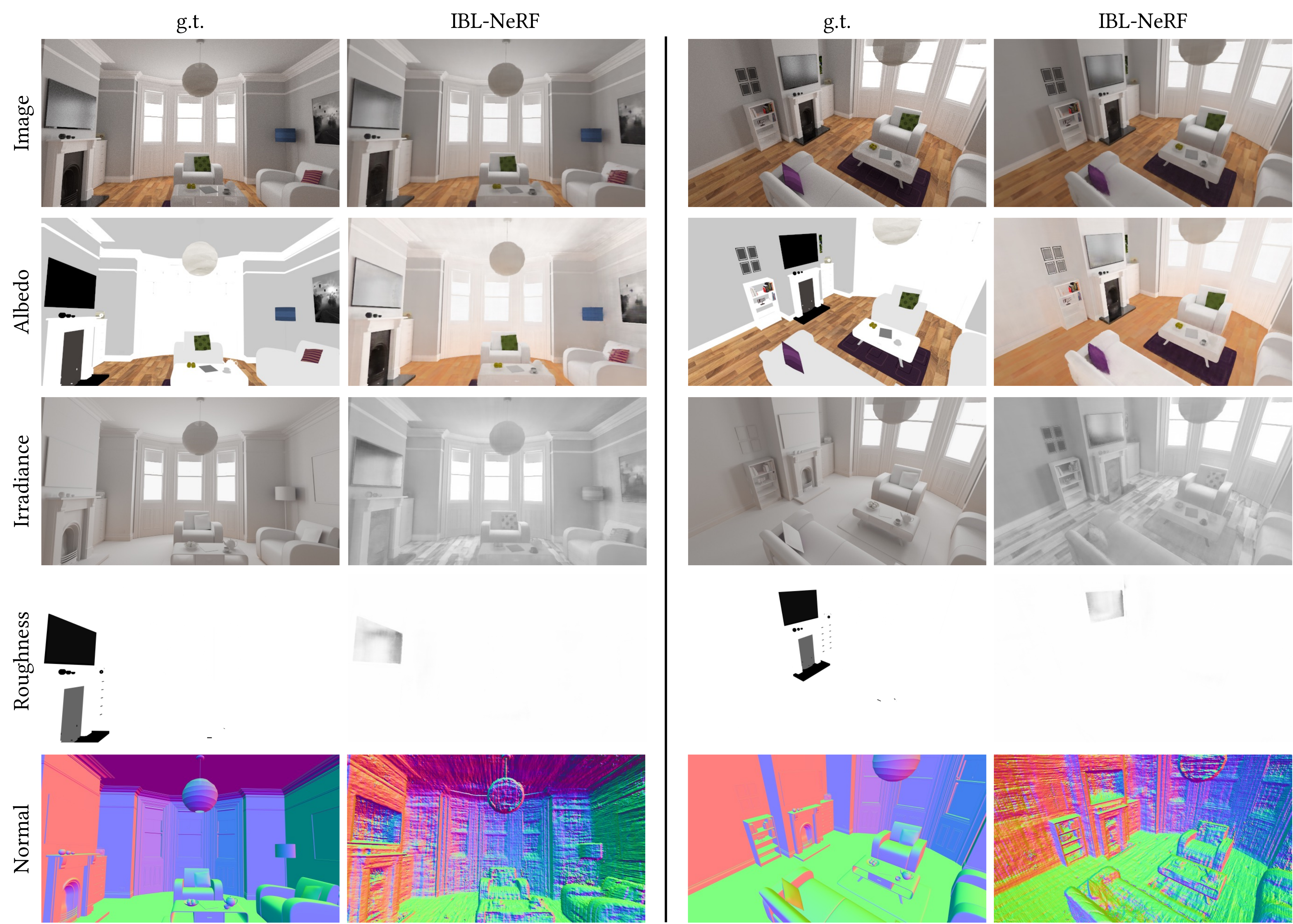}
    \caption{Additional qualitative results of novel view image synthesis and intrinsic decomposition in \textsc{Livingroom2}.}
    \label{fig:additional_livingroom}
\end{figure*}

\begin{figure*}
    \centering
    \includegraphics[width=\linewidth]{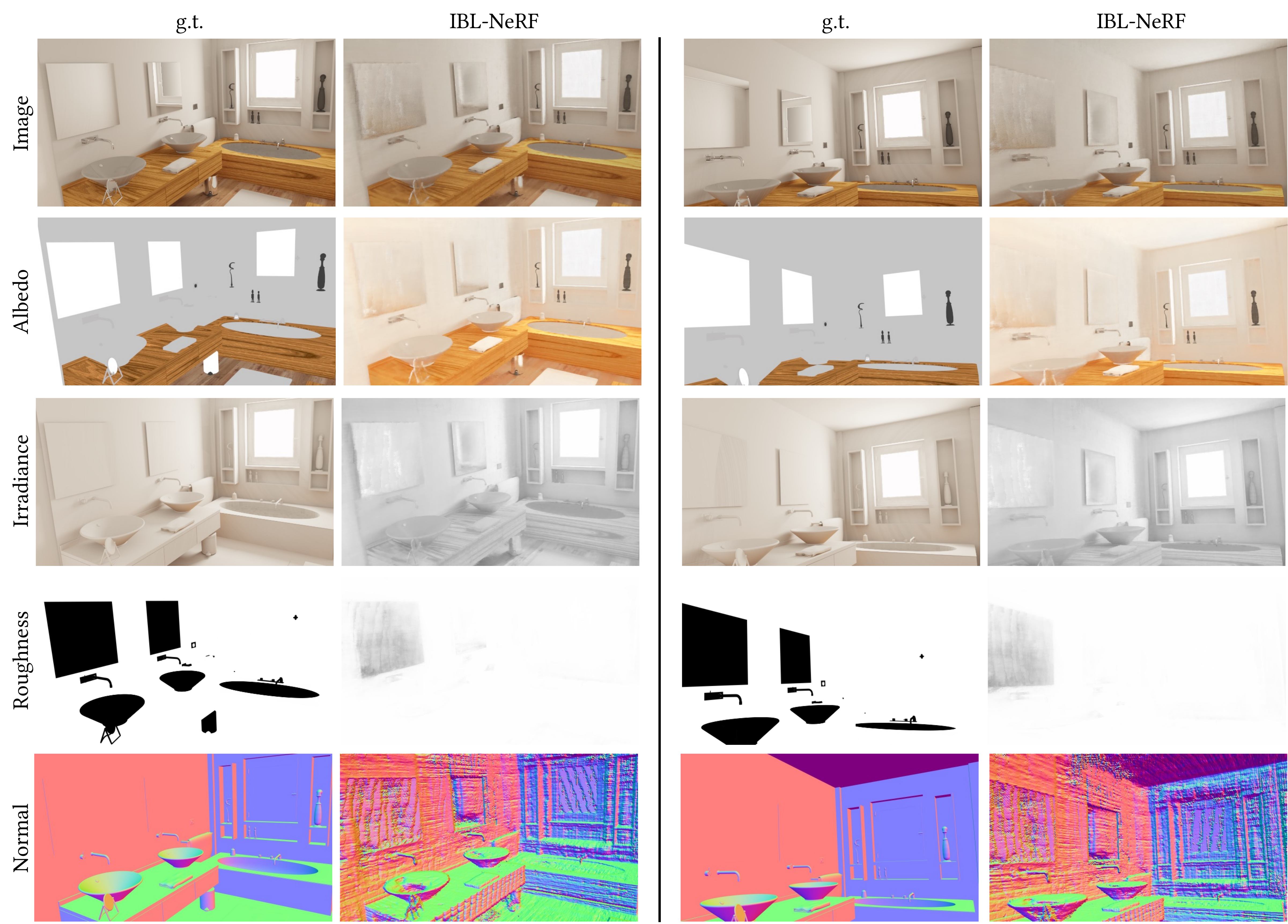}
    \caption{Additional qualitative results of novel view image synthesis and intrinsic decomposition in \textsc{Bathroom2}.}
    \label{fig:additional_bathroom}
\end{figure*}

\begin{figure*}
    \centering
    \includegraphics[width=\linewidth]{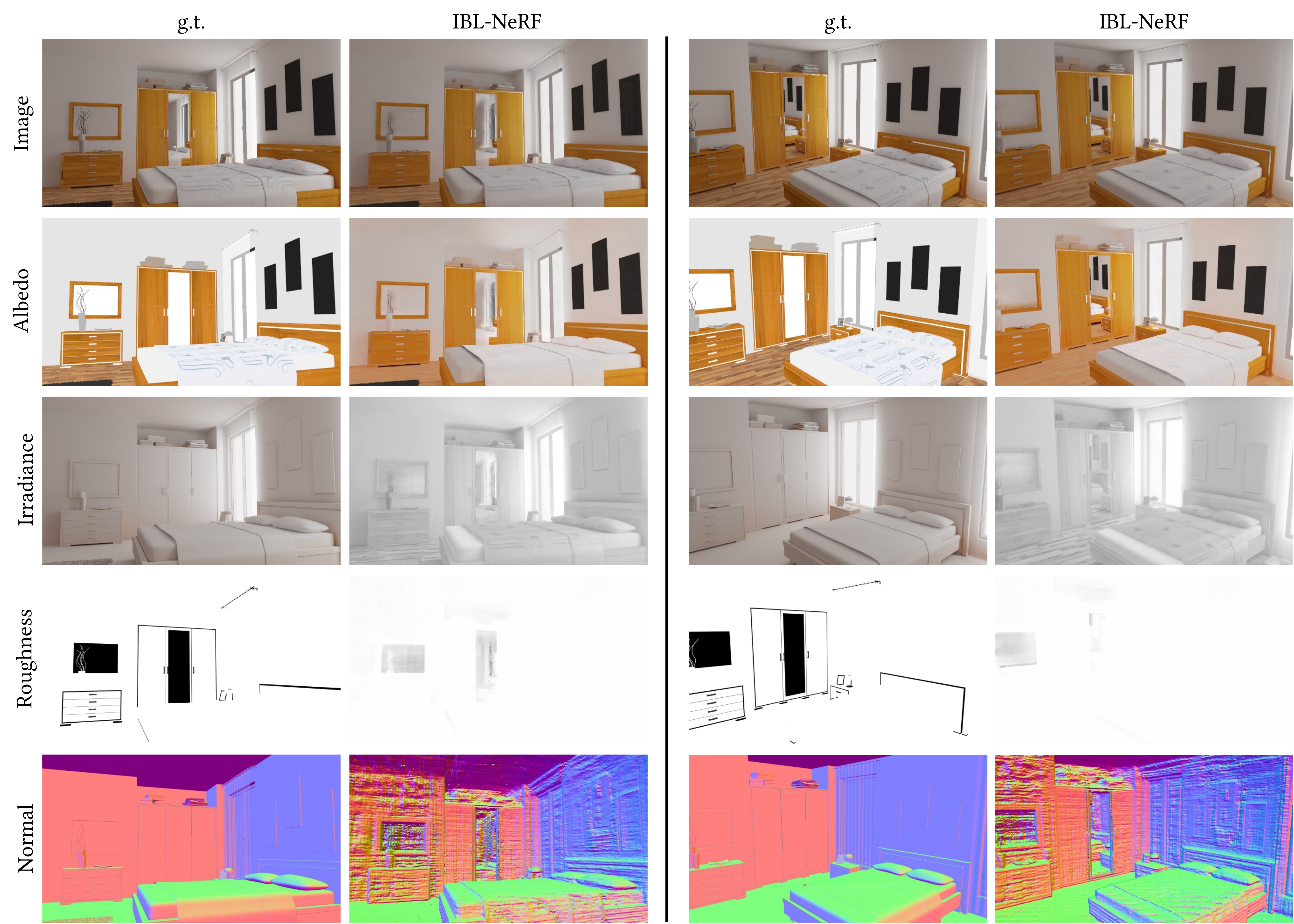}
    \caption{Additional qualitative results of novel view image synthesis and intrinsic decomposition in \textsc{Bedroom}.}
    \label{fig:additional_bedroom}
\end{figure*}

In addition to Fig.~\ref{fig:intrinsic_decomposition_kitchen} and Fig.~\ref{fig:intrinsic_decomposition_livingroom} in the main manuscript, we display more quantitative results of novel view synthesis and intrinsic decomposition in Fig.~\ref{fig:additional_kitchen},~\ref{fig:additional_livingroom}, ~\ref{fig:additional_bathroom} and ~\ref{fig:additional_bedroom}.

\newpage
